\documentclass[sigplan,10pt]{acmart}

\usepackage{amsmath,amsfonts}
\usepackage{algorithmic}
\usepackage{algorithm}
\renewcommand{\algorithmicrequire}{\textbf{Input:}}
\renewcommand{\algorithmicensure}{\textbf{Output:}}
\usepackage{array}
\usepackage{textcomp}
\usepackage{stfloats}
\usepackage{url}
\usepackage{verbatim}
\usepackage{graphicx}

\usepackage{tikz}
\usepackage{enumitem}

\usepackage{amsmath}
\usepackage{amsthm}
\usepackage{mathtools}
\usepackage{tabularx}

\usepackage{pgfplots}
\pgfplotscreateplotcyclelist{mycolors}{
    blue,every mark/.append style={fill=blue},mark=*\\
    red,every mark/.append style={fill=red},mark=square*\\
    brown,every mark/.append style={fill=brown},mark=otimes*\\
    teal,every mark/.append style={fill=teal},mark=diamond*\\
    orange,every mark/.append style={fill=orange},mark=*\\
    cyan,every mark/.append style={fill=cyan},mark=square*\\
    purple,every mark/.append style={fill=purple},mark=otimes*\\
    olive,every mark/.append style={fill=olive},mark=diamond*\\
}
\usepackage{xfrac}
\usepackage{color}

\usepackage{commath}
\usepackage{amsmath}
\usepackage{pgfplots}
\usetikzlibrary{patterns}

\usepackage{xfrac}
\usepackage{amsfonts}
\usepackage{enumitem}
\usepackage{physics}

\usepackage{todonotes}
\usepackage{adjustbox}
\usepackage{multirow}
\usepackage{makecell}
\usepackage{filecontents}

\usepackage{pgfplots}
\usepackage{subcaption}
\usepackage{url}
\usepackage{hyperref}

\usepackage{xcolor} 

\usetikzlibrary{shadows}
\usetikzlibrary{shapes.geometric}
\usepgfplotslibrary{fillbetween}
\usepgfplotslibrary{groupplots}
\pgfplotsset{compat=1.16}
\usetikzlibrary{positioning}
\usepackage{tcolorbox}

\usepackage[capitalize,noabbrev]{cleveref}

\newcommand{\accordion}{\texttt{\textsc{NeuroFlux}}}
\renewcommand\footnotetextcopyrightpermission[1]{}

\definecolor{mutedteal}{RGB}{190,230,230}

\usetikzlibrary{decorations.text}

\AtBeginDocument{%
  }

\copyrightyear{2024}
\acmYear{2024}
\acmConference[Accepted to EuroSys '24]{}{April 22--25, 2024}{Athens, Greece}
\acmBooktitle{EuroSys '24}

\begin{document}

\title{\texttt{{\accordion{}}}: Memory-Efficient CNN Training Using Adaptive Local Learning}

\author{Dhananjay Saikumar}
\affiliation{%
  \institution{University of St Andrews, UK}
  \country{}
}

\author{Blesson Varghese}
\affiliation{%
  \institution{University of St Andrews, UK}
  \country{}
}



\renewcommand{\shortauthors}{Saikumar et al.}

\begin{abstract}
Efficient on-device Convolutional Neural Network (CNN) training in resource-constrained mobile and edge environments is an open challenge. Backpropagation is the standard approach adopted, but it is GPU memory intensive due to its strong inter-layer dependencies that demand intermediate activations across the entire CNN model to be retained in GPU memory. This necessitates smaller batch sizes to make training possible within the available GPU memory budget, but in turn, results in substantially high and impractical training time. We introduce \accordion{}, a novel CNN training system tailored for memory-constrained scenarios. We develop two novel opportunities: firstly, adaptive auxiliary networks that employ a variable number of filters to reduce GPU memory usage, and secondly, block-specific adaptive batch sizes, which not only cater to the GPU memory constraints but also accelerate the training process. \accordion{} segments a CNN into blocks based on GPU memory usage and further attaches an auxiliary network to each layer in these blocks. This disrupts the typical layer dependencies under a new training paradigm - \textit{`adaptive local learning'}. Moreover, \accordion{} adeptly caches intermediate activations, eliminating redundant forward passes over previously trained blocks, further accelerating the training process. The results are twofold when compared to Backpropagation: on various hardware platforms, \accordion{} demonstrates training speed-ups of 2.3$\times$ to 6.1$\times$ under stringent GPU memory budgets, and \accordion{} generates streamlined models that have 10.9$\times$ to 29.4$\times$ fewer parameters. 
\end{abstract}





\maketitle

\section{Introduction}
\label{sec:introduction}
Convolutional Neural Networks (CNNs) underpin a wide range of applications that run on resource-constrained environments, such as mobile and edge computing systems. These applications may be functionality-related or mission-critical and are required to operate within fixed resource constraints. Examples of the former type of applications include facial authentication~\cite{face_Luttrell, face_Luttrell2}, speech recognition~\cite{REFL, Speech_1, Speech_2, Speech_3}, and gesture detection~\cite{s22030706, Guo_gesture} on home gadgets, and of the latter include advanced perception~\cite{perception, Wu_2017_CVPR_Workshops} and decision-making~\cite{decision_making, HAPI} for autonomous vehicles or robots. 

High-throughput and budget-friendly inference of CNNs comprising millions of parameters has become feasible on mobile and edge due to two recent advances~\cite{mobilegpu, mobilegpu2}. Firstly, the availability of embedded hardware accelerators, such as Graphics Processing Units (GPUs) and Neural Processing Units (NPUs). Secondly, the development of techniques that reduce the complexity of CNNs via pruning~\cite{Hanprune, lth, synflow}, quantization~\cite{quant1}, knowledge distillation~\cite{Knowledge_Distillation}, and neural architecture search~\cite{NAS1, NAS2}. 

However, \textbf{CNN training is still computationally and memory intensive}, necessitating the reliance on servers found in high-performance computing sites or cloud data centers. The typical approach involves either training the CNN entirely on an external server and then deploying it onto the device solely for inference~\cite{HAPI}, or collaboratively training across both the device and the server by leveraging computational offloading~\cite{fedadapt, splitgp, split1, split2, split4}.

Nonetheless, \textbf{on-device training remains a challenge}. This challenge arises from the memory-intensive nature of the widely used \textit{Backpropagation} (BP) paradigm in CNN training. Within this paradigm, it is imperative that the intermediate outputs, from the layers of the CNN, referred to as activations, are retained in memory for calculating gradients (more in Section~\ref{subsec:backprop}). The activations of deep models require substantial amounts of GPU memory. For instance, a production-quality CNN model, such as ResNet-18, when used in real-world scenarios~\cite{real_world1,real_world2,real_world3} may require up to 15 GB of GPU memory. While limited computing resources may allow for on-device training, albeit at the expense of training time, the lack of memory on mobile and edge resources is simply prohibitive for on-device CNN training. Consequently, on-device training is limited to smaller batch sizes, which in turn increases training time, or mandates the use of simpler models and datasets.
Our work is therefore positioned to achieve a breakthrough in on-device CNN training in resource-constrained environments by developing the \accordion{} system. \textit{We depart from BP and develop an adaptive layer-wise local learning approach}. The key advantage of the system is memory efficiency by: (a) segmenting the CNN into blocks for reducing the memory requirements as we eliminate inter-layer dependencies, and (b) implementing variable batch sizes for different blocks to maximize memory utilization. Our \textbf{research contributions} are:

1) \textbf{\accordion{}, a system designed for efficient on-device CNN training under memory constraints}. When compared to BP, \accordion{} showcases speedups in training ranging from 2.3$\times$ to 6.1$\times$ and 3.3$\times$ to 10.3$\times$ when compared to BP and classic local learning~\cite{LL_Vanilla}, respectively, given a memory budget.

2) \textbf{Two novel adaptive strategies for local learning}. Firstly, \textit{adaptive auxiliary networks} with variable filters per CNN layer, and secondly, \textit{adaptive batch sizes} that tailors batch size per layer. These strategies allow \accordion{} to process larger training batches, consequently reducing the number of gradient descent steps. This reduction directly improves the training latency of \accordion{} and enables it to achieve higher accuracy than BP and classic local learning in a given time frame.

3) \textbf{Block-based learning that groups CNN layers by memory needs}. Only the actively trained block occupies device memory, while the other blocks are located in storage. 
This combined with the adaptive strategies, allows for training to be carried out on a fixed memory budget.

4) \textbf{Streamlined early exit model}. The local learning strategy of \accordion{} selects the optimal early exit model. CNNs generated by \accordion{} are 10.9$\times$ to 29.4$\times$ more compact than those from BP and classic local learning. This naturally impacts inference speed, with \accordion{} models showing throughput improvements of 1.61$\times$ to 3.95$\times$ across different hardware platforms.

The remainder of this paper is organized as follows. 
Section~\ref{sec:background} presents the background and motivation for the research.
Section~\ref{sec:opportunities} presents a case for adaptive local learning that underpins our work. 
Section~\ref{sec:overview} provides an overview of \accordion{}, the system developed to leverage adaptive local learning.
Section~\ref{sec:design} considers the components of \accordion{} and the underlying techniques.
Section~\ref{sec:studies} evaluates \accordion{} and highlights the results obtained.
Section~\ref{sec:related_work} considers related work. 
Section~\ref{sec:conclusions} concludes this paper. 
\section{Background and Motivation}
\label{sec:background}

This section further considers the memory-intensive nature of the Backpropagation paradigm and considers viable alternate training paradigms.  

\subsection{On-device Training}
\label{subsec:On_device_training}

The need for on-device training is rising to enable personalized models and ensure data privacy. On-device training underpins recent training paradigms, such as federated learning~\cite{com1} and continual learning~\cite{PARISI201954} that make use of computational resources on edge devices while safeguarding user privacy. Furthermore, on-device training reduces the reliance on extensive backend infrastructure, which enables services to become more digitally sovereign and scale rapidly while also being less impacted by network variability~\cite{SAGE}. This ensures user data, including sensitive information like voice recordings and facial images, remains on the device~\cite{iMon, Sign_Language}. On-device training is in contrast to traditional and resource-intensive model development methods, such as Neural Architecture Search (NAS) and structured pruning that require substantial computational resources - from several GPU days~\cite{proxylessnas,DNNShifter} for NAS to GPU hours to days~\cite{EasiEdge,molchanov2017pruning} for structured pruning when utilizing server-grade hardware. These require significant energy and computation but also lead to further on-device fine-tuning costs to tailor the derived models for specific tasks. On-device training emerges as a more streamlined and cost-efficient alternative that can reduce the computational and carbon footprint associated with Deep Neural Network (DNN)  training~\cite{Carbon_Footprint, Green_AI}.

\subsection{Backpropagation is Memory Intensive}
\label{subsec:backprop}

Backpropagation (BP)-based training operates in two phases: the forward pass and the backward pass. In the forward pass, the input data sequentially traverses through the layers of the CNN model, producing intermediate activations that are stored in the memory of the accelerator, such as the GPU. The output of the forward pass is fed into a function, such as Cross Entropy or Mean Squared Error, in order to evaluate a global loss. Conversely, in the backward pass, parameter gradients for the final layer are computed using both the loss and the activations from the preceding layer. The gradients are then propagated backwards, repeating this through the model. The gradient of each layer depends on the activations of its preceding layer, which makes it necessary to retain all activations in memory throughout the forward pass of the model. The necessity for retaining these activations is presented in Appendix~\ref{sec:appendix-BP}.

The two phases have computational similarities, with the backward pass requiring up to 3$\times$ the FLOPs (floating-point operations) as the forward pass. The forward pass during inference and training is computationally identical, differing only in GPU memory requirements, as the latter requires all intermediate activations for subsequent gradient calculations in the backward pass. Mobile and edge-based hardware accelerators can meet the compute demands of the forward and backward passes, but satisfying memory requirements is a challenge.

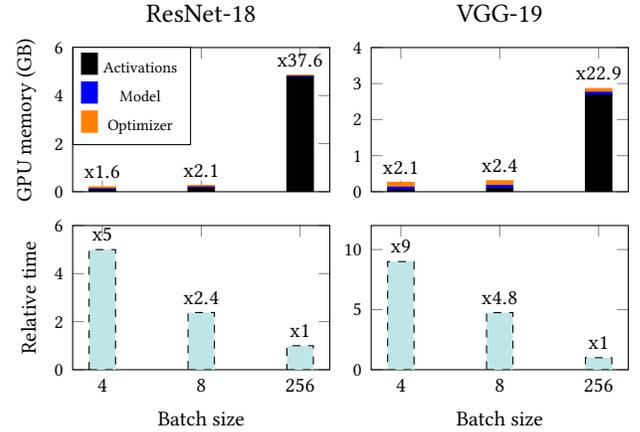
\begin{figure}[t]
\centering
\captionsetup{belowskip=-10pt} 
\begin{tikzpicture}
\begin{groupplot}[
group style={
group size=2 by 2,
horizontal sep=0.55cm,
vertical sep=0.45cm,
},
height=3.5cm,
width=5cm,
ybar stacked,
enlarge x limits=0.15,
ymin=0,
symbolic x coords={4, 8, 256},
yticklabel style={/pgf/number format/fixed, font=\scriptsize},
xticklabel style={font=\footnotesize},
legend columns=1,
legend style={at={(0.23,1.005)}, anchor=north, font=\tiny, /tikz/every even column/.append style={column sep=0.5cm}}
]
\nextgroupplot[title=ResNet-18, ylabel=GPU memory (GB), legend entries={Activations, Model, Optimizer}, xticklabels=\empty, ymax = 6, ylabel style={font=\footnotesize}]
\addplot+[ybar, fill=black, draw=black] coordinates {
    (4, 0.0842)
    (8, 0.1572)
    (256, 4.7629)
};
\addplot+[ybar, fill=blue, draw=blue] coordinates {
    (4, 0.0420)
    (8, 0.0420)
    (256, 0.0420)
};
\addplot+[ybar, fill=orange, draw=orange] coordinates {
    (4, 0.07868)
    (8, 0.06718)
    (256, 0.04419)
};

\pgfmathsetmacro{\multiplierfourR}{(0.0842+0.0420+0.0768)/0.1289}
\pgfmathsetmacro{\multipliereightR}{(0.1572+0.0420+0.0656)/0.1289}
\pgfmathsetmacro{\multipliertwosixR}{(4.7629+0.0420+0.0431)/0.1289}
\node[black, font=\footnotesize, yshift=2mm] at (axis cs:4,0.203) {x\pgfmathprintnumber[precision=1]{\multiplierfourR}};
\node[black, font=\footnotesize, yshift=2mm] at (axis cs:8,0.265) {x\pgfmathprintnumber[precision=1]{\multipliereightR}};
\node[black, font=\footnotesize, yshift=2mm] at (axis cs:256,4.848) {x\pgfmathprintnumber[precision=1]{\multipliertwosixR}};

\nextgroupplot[title=VGG-19, xticklabels=\empty, ymax = 4]
\addplot+[ybar, fill=black, draw=black] coordinates {
    (4, 0.0519)
    (8, 0.0966)
    (256, 2.6838)
};
\addplot+[ybar, fill=blue, draw=blue] coordinates {
    (4, 0.0751)
    (8, 0.0751)
    (256, 0.0751)
};
\addplot+[ybar, fill=orange, draw=orange] coordinates {
    (4, 0.1347)
    (8, 0.1333)
    (256, 0.1042)
};

\pgfmathsetmacro{\multiplierfourV}{(0.0519+0.0751+0.1347)/0.1248}
\pgfmathsetmacro{\multipliereightV}{(0.0966+0.0751+0.1333)/0.1248}
\pgfmathsetmacro{\multipliertwosixV}{(2.6838+0.0751+0.1042)/0.1248}
\node[black, font=\footnotesize, yshift=2mm] at (axis cs:4,0.2617) {x\pgfmathprintnumber[precision=1]{\multiplierfourV}};
\node[black, font=\footnotesize, yshift=2mm] at (axis cs:8,0.305) {x\pgfmathprintnumber[precision=1]{\multipliereightV}};
\node[black, font=\footnotesize, yshift=2mm] at (axis cs:256,2.8631) {x\pgfmathprintnumber[precision=1]{\multipliertwosixV}};

\nextgroupplot[ylabel=Relative time, ymax = 6, ylabel style={font=\footnotesize}, xlabel=Batch size, xlabel style={font=\footnotesize}]
\pgfmathsetmacro{\reltimefourR}{58.33/11.67}
\pgfmathsetmacro{\reltimeeightR}{27.78/11.67}
\addplot+[ybar, fill=mutedteal, draw=black, dashed] coordinates {
    (4, \reltimefourR)
    (8, \reltimeeightR)
    (256, 1)
};

\node[black, font=\footnotesize, yshift=2mm] at (axis cs:4,\reltimefourR) {x\pgfmathprintnumber[precision=1]{\reltimefourR}};
\node[black, font=\footnotesize, yshift=2mm] at (axis cs:8,\reltimeeightR) {x\pgfmathprintnumber[precision=1]{\reltimeeightR}};
\node[black, font=\footnotesize, yshift=2mm] at (axis cs:256,1) {x1};

\nextgroupplot[title={}, ymax = 12, xlabel=Batch size, xlabel style={font=\footnotesize}]
\pgfmathsetmacro{\reltimefourV}{60/6.666}
\pgfmathsetmacro{\reltimeeightV}{31.67/6.666}
\addplot+[ybar, fill=mutedteal, draw=black, dashed] coordinates {
    (4, \reltimefourV)
    (8, \reltimeeightV)
    (256, 1)
};
\node[black, font=\footnotesize, yshift=2mm] at (axis cs:4,\reltimefourV) {x\pgfmathprintnumber[precision=1]{\reltimefourV}};
\node[black, font=\footnotesize, yshift=2mm] at (axis cs:8,\reltimeeightV) {x\pgfmathprintnumber[precision=1]{\reltimeeightV}};
\node[black, font=\footnotesize, yshift=2mm] at (axis cs:256,1) {x1};
\end{groupplot}
\end{tikzpicture}
\caption{Comparison of GPU memory usage and relative training time for different architectures and batch sizes on the Tiny ImageNet dataset. The top row shows memory used by activations, the model, and the optimizer, with multipliers indicating memory required relative to inference. The bottom row highlights training time relative to batch size of 256.}
\label{fig:memory_usage_time}
\end{figure}

In Figure~\ref{fig:memory_usage_time}, we examine the GPU memory consumption and training time of two widely used CNN architectures on the Tiny ImageNet dataset: ResNet-18 and VGG-19. This analysis employs BP-based training across varying batch sizes. The memory footprint comprises three components: the inherent size of the model, the memory overhead of the optimizer, and the memory required for activations. It is immediately evident that the GPU memory usage during training arises from the memory required for retaining the intermediate activations. This poses challenges for training on memory-restricted devices. A similar trend is observed for smaller models specifically designed for mobile environments, such as MobileNet~\cite{MobileNet}, trained on the Tiny ImageNet dataset. For a batch size of 256, MobileNet requires 830MB of GPU memory to retain activations during BP-based training, while inference can be performed under 35MB. Therefore, even using smaller models tailored for mobile devices, the activations still dominate the GPU memory requirement for BP-based training.

While opting for smaller batch sizes can reduce memory demands, it increases the number of Stochastic Gradient Descent (SGD) steps due to more batches being processed. This increase directly prolongs training time. Additionally, the necessity of preprocessing and loading each batch onto the GPU individually introduces overhead and potential I/O bottlenecks, further slowing training. For instance, VGG-19 trained on the Tiny ImageNet with a batch size of 4 takes over nine times longer than with a batch size of 256.

\subsection{Training Paradigms Beyond Backpropagation}
Several paradigms have been developed to address the limitations of BP-based training. Among them, Feedback Alignment, Signal Propagation, and Local Learning are noteworthy. Local Learning is considered further as it will be leveraged in our work given the opportunities it presents (see Section~\ref{sec:opportunities}). The other paradigms are considered in Section~\ref{sec:related_work}.

1) \textbf{\textit{Feedback Alignment}} (FA): FA provides a solution to the `weight transport problem' in DNN training~\cite{FA1, FA2}. While FA is shown to be effective for certain DNN models, it is not ideal for CNNs~\cite{Signal_prop}.

2) \textbf{\textit{Signal Propagation}} (SP): By employing forward passes only, SP offers a unique layer-wise training methodology, akin to LL, but without making use of auxiliary networks~\cite{Signal_prop}. While SP showcases a memory efficiency over BP, its accuracy is lower than BP and LL. Thus, it is not widely adopted for training.

\begin{figure}[t]
  \centering
  \includegraphics[width=0.36\textwidth]{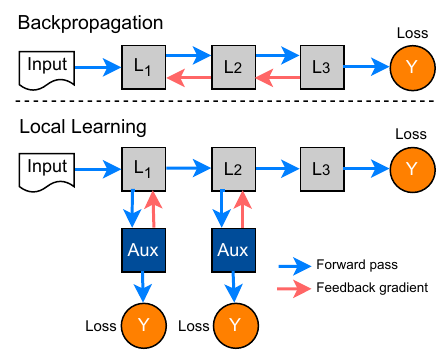}
  \vspace{-10pt}
\caption{Comparison of BP and LL. BP relies on a global loss, with updates for each layer dependent on subsequent layers. In contrast, LL pairs each layer (excluding the last) with an auxiliary network for independent updates using local losses, thereby eliminating backward feedback dependencies.}\label{fig:local_learning}
\end{figure}

3) \textbf{\textit{Local Learning}} (LL) assigns an auxiliary network to each CNN layer to facilitate local prediction, rather than relying on a global loss derived from the final output of the CNN as in BP~\cite{LL_Vanilla} (refer Figure~\ref{fig:local_learning}). When a training batch is provided to the first CNN layer, it is processed to produce output activations. The activations are fed into the auxiliary network for prediction. Based on the prediction and the true labels, a local loss is computed. The loss is subsequently used to optimize the parameters in both the first layer and its auxiliary network. The activations of the first layer are then passed on to the next layer. This is sequential and continues for every layer, ensuring that each layer processes the data and undergoes optimization until the entire batch has traversed all layers~\cite{DGL}. Any feedback dependencies on subsequent layers are eliminated.

Recent empirical studies demonstrate that LL achieves similar performance as BP-based training for a range of datasets~\cite{LL_Vanilla}. LL does not require all intermediate activations to be retained in memory to update parameters.~\cite{LL_Vanilla}, thereby offering the potential for memory-efficient training as the parameters are updated on a layer—by—layer basis (layer-wise training). However, in the existing work on classic LL, this memory gain is offset by the memory requirements of the auxiliary network to make predictions. The opportunities we will leverage within this training paradigm are considered in the next section.

\begin{figure}[t]
\centering
\begin{tikzpicture}
    \begin{axis}[
        height=4.5cm,
        xlabel={GPU memory required},
        ylabel={Accuracy},
        xmin=-10, xmax=10,
        ymin=-10, ymax=10,
        xtick={-10,0,10},
        ytick={-10,0,10},
        xticklabels={Low,,High},
        yticklabels={Low,,High},
        scatter/classes={
            BP={semi-mutedbrown},
            LL={semi-mutedolive},
            FA={semi-mutedgray},
            SP={semi-mutedorange}
        },
        font=\footnotesize,
        legend style={at={(0.5,-0.35)},anchor=north}, 
        legend columns=-1, 
        axis lines=middle, 
        enlarge y limits=false,
        enlarge x limits=false,
        every axis plot post/.style={mark options={solid}},
        xlabel style={at={(axis description cs:0.5,-0.1)},anchor=north}, 
        ylabel style={at={(axis description cs:-0.15,0.5)},anchor=south, rotate=90} 
    ]

    \fill[blue!15] (axis cs:-10,9) rectangle (axis cs:0,0);

    \definecolor{semi-mutedbrown}{RGB}{175,90,15}
    \definecolor{semi-mutedolive}{RGB}{128,140,67}
    \definecolor{semi-mutedgray}{RGB}{150,150,150}
    \definecolor{semi-mutedorange}{RGB}{255,165,45}

    \addplot[scatter, only marks, scatter src=explicit symbolic, mark size=3pt]
        coordinates {
            (7,9) [BP]
            (9,9) [LL]
            (-3,-3) [FA]
            (-5,-2) [SP]
        };
    \legend{BP, LL, FA, SP}
    \end{axis}
\end{tikzpicture}
\vspace{-6pt}
\caption{GPU memory required and accuracy achieved by different training paradigms. The blue-shaded quadrant represents the ideal zone for a training paradigm (low GPU memory utilization and high accuracy).}
\label{fig:comparison}
\end{figure}
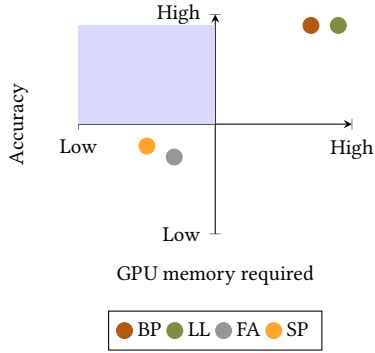

\textbf{Comparing GPU memory vs accuracy}: 
Figure~\ref{fig:comparison} captures the GPU memory required and accuracy of the training paradigms. 
An ideal paradigm will achieve high accuracy without significant memory use. 
Although LL achieves comparable accuracy to BP, it still has substantial memory requirements. 
\textit{Currently there are no training paradigms that can achieve high accuracy with a low GPU memory utilization}.
Therefore, we explore the potential to lower the memory utilization of LL into the blue-shaded area of Figure~\ref{fig:comparison}.

\section{The Case for Adaptive Local Learning}
\label{sec:opportunities}
We have identified \textbf{two opportunities} for reducing the GPU memory requirements while achieving comparable accuracy to BP. Here we introduce the notion of \textit{`adaptivity'} for the first time within local learning that is considered below. 

\subsubsection*{Opportunity 1: Adaptive Auxiliary Networks in Local Learning}
In LL, each CNN layer (except the final layer) is accompanied by an auxiliary network, as shown in Figure~\ref{fig:local_learning}. Classic LL employs a CNN classifier as the auxiliary network, which comprises a convolution layer, average pooling, and a fully-connected layer for prediction~\cite{LL_Vanilla}. 

In the classic LL approach, auxiliary networks use a fixed number of convolutional filters (256 filters)~\cite{LL_Vanilla}. However, the GPU memory used during classic LL training is noted to be higher than BP, although in LL, only the activations of a single layer and its corresponding auxiliary network are retained in GPU memory. This is due to the number of convolutional filters in the auxiliary networks of the initial layers of the CNN, resulting in large activations that are not sufficiently downsampled. 
Decreasing the number of convolutional filters uniformly for all auxiliary networks of the CNN leads to reduced GPU memory usage. However, the accuracy obtained is much lower than BP. 

Instead, we develop a new strategy referred to as \textbf{`Adaptive Auxiliary Networks based LL (AAN-LL)’} - employing a variable number of convolutional filters for the auxiliary networks for achieving both lower GPU memory consumption and comparable accuracy to BP. For the initial layers of the CNN, which in this context are defined as the layers before the first downsampling operation, the number of convolutional filters in the auxiliary network is halved with respect to the CNN layer with the smallest number of convolutional filters. For example, the smallest number of convolutional filters for a VGG model is 64; the number of convolutional filters of the auxiliary network for the initial layers will therefore be 32. For the subsequent CNN layers, we reduce the number of convolutional filters again by half with respect to the CNN layer with the largest number of convolutional filters. For example, the largest number of convolutional filters for a VGG model is 512; therefore, the number of convolutional filters of the auxiliary network for the later layers is 256. 

\begin{figure}[t]
    \centering
    \captionsetup{belowskip=-5pt} 
    \begin{tikzpicture}
        \begin{axis}[
            width=0.45\textwidth,
            height=0.22\textwidth, 
            xlabel={Batch size},
            ylabel={GPU memory (MB)},
            xmin=10, xmax=95,
            xtick={10,30,...,90},
            ymin=0, ymax=6000,
            legend pos=north west,
            legend style={font=\footnotesize, cells={anchor=west}, inner sep=0.1pt,legend columns=2},
            tick label style={font=\footnotesize},
            label style={font=\footnotesize},
            legend cell align=left,
        ]

        \addplot[color=brown,mark=square*,line width=1pt] coordinates {
            (1,232.448) 
            (6,393.216) 
            (11,605.184) 
            (16,816.128) 
            (21,1029.12) 
            (26,1235.072) 
            (31,1449.984) 
            (36,1664.896) 
            (41,1877.952) 
            (46,2087.936) 
            (51,2302.976) 
            (56,2516.96) 
            (61,2729.872) 
            (66,2944.832) 
            (71,3160.96) 
            (76,3368.064) 
            (81,3589.056) 
            (86,3800.928) 
            (91,4018.624) 
            (96,4231.744)
        };
        \addlegendentry{BP}

        \addplot[color=gray,dashed,line width=1pt] coordinates {
            (1,91.136) 
            (6,152.576) 
            (11,214.016) 
            (16,273.408) 
            (21,334.848) 
            (26,395.264) 
            (31,456.704) 
            (36,517.12) 
            (41,578.56) 
            (46,638.976) 
            (51,699.392) 
            (56,760.832) 
            (61,822.272) 
            (66,883.712) 
            (71,945.152) 
            (76,1005.568) 
            (81,1066.944) 
            (86,1127.04) 
            (91,1188.9856) 
            (96,1249.1584)
        };
        \addlegendentry{Inference}

        \addplot[color=olive,mark=triangle*,line width=1pt] coordinates {
            (1,184.32) 
            (6,464.192) 
            (11,744.448) 
            (16,1025.728) 
            (21,1306.176) 
            (26,1587.52) 
            (31,1870.464) 
            (36,2149.376) 
            (41,2430.656) 
            (46,2710.112) 
            (51,2990.944) 
            (56,3273.216) 
            (61,3554.624) 
            (66,3836.896) 
            (71,4119.424) 
            (76,4398.88) 
            (81,4681.344) 
            (86,4960.896) 
            (91,5239.68) 
            (96,5518.112)
        };
        \addlegendentry{Classic LL}

        \addplot[color=teal,mark=pentagon*,line width=1pt] coordinates {
            (1,152.576) 
            (6,250.88) 
            (11,355.328) 
            (16,460.8) 
            (21,565.248) 
            (26,670.72) 
            (31,776.192) 
            (36,881.664) 
            (41,987.136) 
            (46,1091.52) 
            (51,1196.992) 
            (56,1301.28) 
            (61,1406.752) 
            (66,1511.2) 
            (71,1616.672) 
            (76,1721.12) 
            (81,1826.592) 
            (86,1930.368) 
            (91,2034.688) 
            (96,2138.816)
        };
        \addlegendentry{AAN-LL}
        \end{axis}
    \end{tikzpicture}
    \vspace{-8pt}
\caption{GPU memory usage of VGG-19 for inference, Backpropagation (BP), classic Local Learning (LL) with a constant number of 256 convolutional filters~\cite{LL_Vanilla} and the proposed Adaptive Auxiliary Networks-based LL (AAN-LL) for different batch sizes.}
    \label{fig:peak_memory_comparison}
\end{figure}
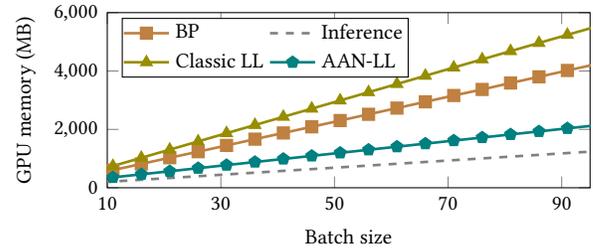

The GPU memory trends for inference, BP, classic LL with 256 convolutional filters and our strategy AAN-LL are shown in Figure~\ref{fig:peak_memory_comparison} for VGG-19; similar trends are observed for ResNet-18.
The success of the AAN-LL strategy is in reducing the size of the activations for the initial layers to reduce GPU memory consumption, while maintaining a sufficiently large number of convolutional filters in the later layers to ensure effective learning. 

\subsubsection*{Opportunity 2: Adaptive Batch Sizes in Local Learning}
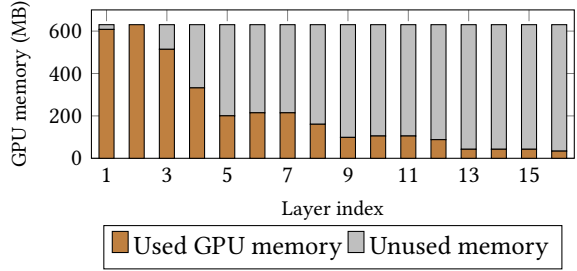
\begin{figure}[t]
    \centering
    \captionsetup{belowskip=-10pt} 
    \begin{tikzpicture}
        \begin{axis}[
            width=0.45\textwidth,
            height=0.2\textwidth, 
            ybar stacked, 
            bar width=0.2cm,
            ylabel={GPU memory (MB)},
            xlabel={Layer index},
            xtick={1,3,5,7,9,11,13,15},
            xticklabels={1,3,5,7,9,11,13,15},
            tick label style={font=\small},
            label style={font=\footnotesize},
            enlarge x limits={abs=0.5},
            ymin=0,
            ymax=700,
            legend style={at={(0.5,-0.45)}, anchor=north, legend columns=-1},
        ]
        \addplot[fill=brown] coordinates {
            (1,607.97) (2,630.34) (3,514.15) (4,332.14) (5,200.39)
            (6,214.64) (7,214.64) (8,160.64) (9,98.65) (10,105.15)
            (11,105.15) (12,87.9) (13,42.65) (14,42.65) (15,42.65)
            (16,33.83)
        };
        \addplot[fill=gray!50] coordinates { 
            (1,630.34-607.97) (2,0) (3,630.34-514.15) (4,630.34-332.14) (5,630.34-200.39)
            (6,630.34-214.64) (7,630.34-214.64) (8,630.34-160.64) (9,630.34-98.65) (10,630.34-105.15)
            (11,630.34-105.15) (12,630.34-87.9) (13,630.34-42.65) (14,630.34-42.65) (15,630.34-42.65)
            (16,630.34-33.83)
        };
        \legend{Used GPU memory, Unused memory}
        \end{axis}
    \end{tikzpicture}
    \vspace{-6pt}
    \caption{GPU memory usage for training VGG-19 with a batch size of 30 images using AAN-LL. `Unused Memory' area refers to GPU memory not utilized by each layer.}
    \label{fig:layer_decomposition}
\end{figure}

Consider an example - the GPU memory using AAN-LL is approximately 630MB for training VGG-19 using a batch size of 30; refer to Figure~\ref{fig:peak_memory_comparison}. The GPU memory use of AAN-LL arises from the initial layers of the CNN. The activations are gradually downsampled as they progress through the model, resulting in larger activations for the initial layers compared to subsequent ones. Figure~\ref{fig:layer_decomposition} highlights this observation; the second layer of the CNN dominates GPU memory. Thus, training the initial layers of a CNN is a GPU memory bottleneck. They also mandate a small batch size to be used for training the entire CNN under GPU memory constraints. 

\begin{figure}[t]
    \centering
    \captionsetup{belowskip=-10pt} 
    \begin{tikzpicture}
        \begin{axis}[
            width=0.45\textwidth,
            height=0.2\textwidth, 
            ybar,
            bar width=0.2cm,
            ylabel={Maximum batch size},
            xlabel={Layer index},
            xtick={1,3,5,7,9,11,13,15},
            xticklabels={1,3,5,7,9,11,13,15},
            tick label style={font=\small},
            label style={font=\footnotesize},
            enlarge x limits={abs=0.5}, 
            ymin=0,
            ymax=1100 
        ]
        \addplot[fill=brown]  coordinates {
            (1,31) (2,30) (3,37) (4,60) (5,95) 
            (6,90) (7,90) (8,120) (9,195) (10,195) 
            (11,195) (12,240) (13,810) (14,810) (15,810) 
            (16,1000)
        };
        \end{axis}
    \end{tikzpicture}
    \vspace{-10pt}
    \caption{Maximum possible batch sizes per layer when training VGG-19 using AAN-LL while staying below a 630MB GPU memory limit.}
    \label{fig:max_batch_size}
\end{figure}
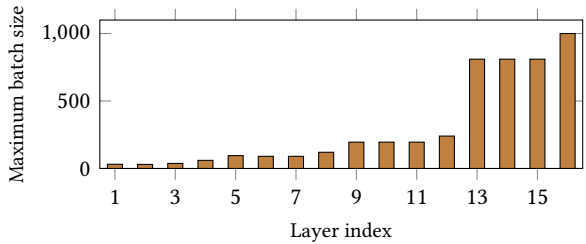

Due to the inherent design of BP and classic LL, they use a fixed batch size for training the entire CNN. However, we make two key observations. Firstly, the later layers of the CNN do not reach the GPU memory use of the initial layers, as shown in Figure~\ref{fig:layer_decomposition}. Secondly, the maximum possible batch size for a given GPU memory budget (630 MB in the case of VGG-19 as it is the maximum GPU memory used) is higher for later CNN layers than the initial layers as shown in Figure~\ref{fig:max_batch_size}. 

We leverage these observations to develop \textbf{`Adaptive Batch-based LL (AB—LL)'}. Contrary to conventional methods that use a static batch size, AB—LL employs dynamic batch sizes for training different layers. While the underlying algorithm for (LL) uses a fixed batch size for training the entire CNN, AB—LL dynamically adjusts the number of activations in a training batch in real-time. This adjustment effectively changes the number of batches consumed by each layer or groups of layers with similar GPU memory requirements. By capitalizing the unused GPU memory of the later CNN layers (shown as `Unused Memory' in Figure~\ref{fig:layer_decomposition}), AB—LL enables the use of larger batch sizes specifically for these layers, thereby optimizing GPU memory use and accelerating training. Beyond mere efficiency enhancements, this approach offers a strategic shift in training by accommodating the fine-grained GPU memory requirements of individual CNN layers or their groups to maximize the utilization of available GPU memory resources.

\section{\accordion{} Overview}
\label{sec:overview}
\begin{figure*}[t]
  \centering
  \includegraphics[width=\textwidth]{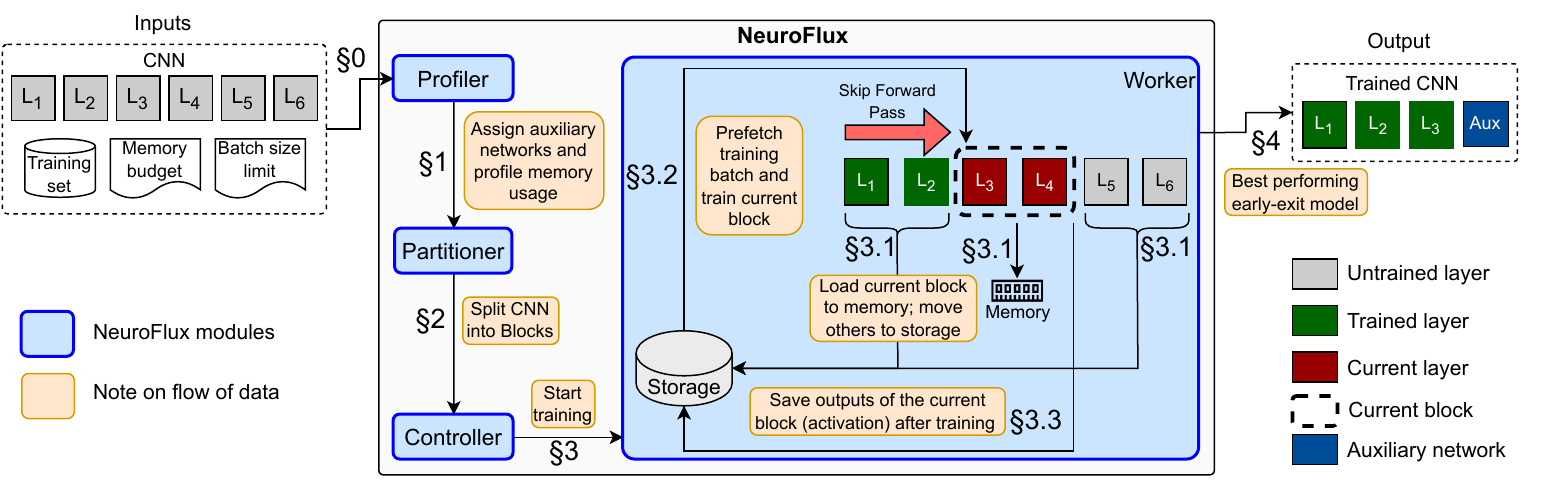}
  \captionsetup{justification=centering, width=\textwidth}
  \caption{The \accordion{} architecture.}
  \label{fig:accordion}
\end{figure*}

We develop \accordion{}, \textit{a CNN training system that leverages adaptive local learning by incorporating adaptive auxiliary networks and adaptive batch sizes within LL to facilitate GPU memory-efficient on-device training in resource-constrained edge environments}. \accordion{} can be used to train large production quality CNNs on smaller devices under memory budgets that cannot be achieved using conventional BP. 

The architecture of \accordion{} is shown in Figure~\ref{fig:accordion} and the individual modules are considered further in the next section. \accordion{} takes four inputs (§0): an untrained CNN, a training set, a GPU memory budget, and a batch size limit. The high-level workflow of \accordion{} is considered next.

\textbf{\textit{Profiler}}: The Profiler first assigns auxiliary networks to the layers of an input CNN by utilizing the AAN-LL strategy presented in the previous section. Following this allocation, the GPU memory utilization required for training each layer is benchmarked for different batch sizes. These measurements subsequently facilitate the construction of layer-specific linear models for predicting GPU memory consumption based on the batch size (§1).
    
\textbf{\textit{Partitioner}}: After obtaining the layer-specific linear models, the Partitioner systematically segments the layers of the given CNN into `blocks' based on their projected GPU memory consumption patterns. Each block consists of contiguous layers with similar GPU memory footprints. Concurrently, the Partitioner determines the optimal training batch size for each block, ensuring the overall training remains within the stipulated GPU memory budget set by the user (§2). 

\textbf{\textit{Controller} and \textit{Worker}}: After receiving the blocks and their respective training batch sizes, the Controller activates CNN training by deploying the Worker (§3). The Worker loads the current block into GPU memory (§3.1). Subsequently, it trains the layers within this block via LL using the pre-determined batch size for the given block (§3.2). Once the block is trained, the activations of its final layer are saved to a storage device (§3.3; they are not stored in GPU memory). After training, the current block is moved to storage, and the subsequent block is loaded into GPU memory (§3.1). The saved activations are inputs for the subsequent block, ensuring continuity in learning, and eliminating the need for forward passes over the trained block(s). This subsequently reduces the compute requirements, which accelerates the overall CNN training. 

Additionally, \accordion{} uses a prefetching mechanism to adjust the number of activations per batch on the fly, such that it has the same batch size recommended by the Partitioner (§3.2). This facilitates the training of different blocks with varied batch sizes via the AB-LL technique, eliminating the need for an entire CNN to be trained using small batch sizes due to the GPU memory demands arising from the initial layers. This process of blockwise training is repeated until all blocks have been trained.

\textbf{Compact Trained CNN}: The concept of early exits in CNNs allows for inferences to be made before reaching the final layer of a model by adding exit points selectively before the final layers~\cite{HAPI, SDN, BranchyNet, 9010043, huang2018multiscale}. These are known to offer higher inference efficiency with low latencies. In the context of \accordion{}, the AAN-LL technique ensures that each CNN layer is paired with its corresponding auxiliary network, which effectively introduces a distinct prediction point for every layer. Following training, \accordion{} systematically evaluates the performance of each layer by taking into account both accuracy and the number of parameters in a given early exit model. Finally, a model with an early exit point that balances the need for achieving high accuracy and has a small resource footprint is generated (§4).

\section{Design}
\label{sec:design}
We first present the mathematical description of the LL paradigm. Then, we discuss the modules and the techniques employed by \accordion{}.

\subsection{Representing Local Learning}
We denote a CNN of \(N\) layers as \(F(\theta)\), where \(\theta\) represents the set of parameters \((\theta_{1},\theta_{2},....,\theta_{N})\). \(\theta_{n}\) represents the parameters of the \(n^{th}\) layer. The CNN architecture integrates the rectified linear unit (ReLU) non-linearity and a down-sampling operation, symbolised by \(\alpha\) and \(P_{n}\), respectively. An input training batch \(x_{0}\) undergoes transformations across the layers of the CNN, resulting in an output, \(x_{n+1}\) at layer \(n\). Subsequently, \(x_{n+1}\) is channelled into an auxiliary network, denoted as \(A_{n}\), which generates an early prediction for the given depth, \(z_{n+1}\). The transformations can be represented mathematically as:
\begin{equation}
\label{eq:1}
\left\{
\begin{aligned}
& x_{n+1} = \alpha P_{n} \theta_{n} x_{n}\\
& z_{n+1} = A_{n} x_{n+1}
\end{aligned}
\right.
\end{equation}

The auxiliary network, $A_{n}$, is a CNN classifier, consisting of a convolution layer characterized by $\beta_{n}$, followed by a downsampling mechanism, $F_{n}$, such as max pooling or average pooling, followed by linear layer(s) $\gamma_{n}$ for prediction. The auxiliary network carries out the following early prediction:
\begin{equation}
\label{eq:2}
\begin{aligned}
A_{n} x_{n+1} = \gamma_{n} F_{n} \beta_{n} x_{n+1}\\
\end{aligned}
\end{equation}

The objective of training at depth $n$ is based on refining parameters $\{\theta_{n}, \beta_{n}, \gamma_{n}\}$ by solving an auxiliary optimization problem. Given a dataset $\{x^{k},y^{k}\}_{k<K}$, and a loss function $\ell$ (e.g., Mean Squared Error or Cross Entropy Error), the goal is to minimize the loss: 
\begin{equation}
\label{eq:3}
\begin{aligned}
\mathcal{L}(z_{n+1};\theta_{n},\beta_{n}, \gamma_{n}) = \frac{1}{K} \sum_{k} \ell(z(x^{k}; \theta_{n}, \beta_{n}, \gamma_{n}),y^{k}) \
\end{aligned}
\end{equation}

After the parameters of a layer are updated through the auxiliary optimization problem, the activations \(x_{n+1}\) serve as the input to the subsequent layer \(n+1\). This process of layerwise optimization continues sequentially until the input training batch progresses through the final layer. 

\subsection{Modules}
The integrated working of the following modules, namely the \textit{Profiler}, \textit{Partitioner}, \textit{Controller}, and \textit{Worker} enable adaptive local learning in \accordion{}. 

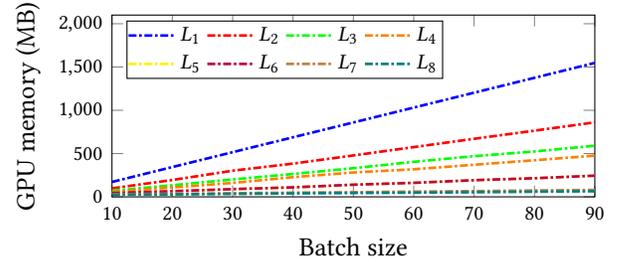
\begin{figure}
    \centering
    \captionsetup{belowskip=-10pt} 
    \begin{tikzpicture}
        \begin{axis}[
            width=0.45\textwidth,
            height=4cm,
            xlabel={Batch size},
            ymin=0, ymax=2100, 
            xmin=10, xmax=90,
            xtick={10,20,...,90},
            ylabel={GPU memory (MB)},
            tick label style={font=\footnotesize},
            legend pos=north west,
            legend style={font=\footnotesize, cells={anchor=west}, legend columns=4,inner sep=0.1pt},
        ]

        \addplot[color=blue, no markers, line width=1pt,densely dashdotted] coordinates {
            (10,172.032) (20,344.064) (30,516.096) (40,688.128) (50,859.136) (60,1031.168) (70,1203.200) (80,1375.232) (90,1547.264)
        };
        \addlegendentry{$L_{1}$}

        \addplot[color=red, no markers, line width=1pt,densely dashdotted] coordinates {
            (10,98.304) (20,194.560) (30,302.080) (40,384.000) (50,478.208) (60,574.464) (70,670.720) (80,764.416) (90,860.672)
        };
        \addlegendentry{$L_{2}$}

        \addplot[color=green, no markers, line width=1pt,densely dashdotted] coordinates {
            (10,69.632) (20,134.144) (30,200.704) (40,264.192) (50,330.752) (60,404.480) (70,470.016) (80,524.288) (90,590.848)
        };
        \addlegendentry{$L_{3}$}

        \addplot[color=orange, no markers, line width=1pt,densely dashdotted] coordinates {
            (10,55.296) (20,108.544) (30,160.768) (40,227.328) (50,281.600) (60,318.464) (70,370.688) (80,422.912) (90,476.160)
        };
        \addlegendentry{$L_{4}$}

        \addplot[color=yellow, no markers, line width=1pt,densely dashdotted] coordinates {
            (10,45.056) (20,65.536) (30,89.088) (40,109.568) (50,140.288) (60,161.792) (70,192.512) (80,215.040) (90,244.736)
        };
        \addlegendentry{$L_{5}$}

        \addplot[color=purple, no markers, line width=1pt,densely dashdotted] coordinates {
            (10,45.056) (20,65.536) (30,89.088) (40,109.568) (50,140.288) (60,161.792) (70,192.512) (80,215.040) (90,244.736)
        };
        \addlegendentry{$L_{6}$}

        \addplot[color=brown, no markers, line width=1pt,densely dashdotted] coordinates {
            (10,30.720) (20,34.816) (30,41.984) (40,48.128) (50,54.272) (60,60.416) (70,66.560) (80,73.728) (90,79.872)
        };
        \addlegendentry{$L_{7}$}

        \addplot[color=teal, no markers, line width=1pt,densely dashdotted] coordinates {
            (10,24.576) (20,27.648) (30,33.792) (40,39.936) (50,44.032) (60,48.128) (70,53.248) (80,59.392) (90,64.512)
        };
        \addlegendentry{$L_{8}$}

        \end{axis}
    \end{tikzpicture}
    \caption{GPU memory required by different layers during training using AAN-LL for varying batch sizes on VGG-11.}
    \label{fig:linear_model}
\end{figure}

\textbf{Profiler}: 
Given an input CNN, a training set, a GPU memory budget, and the batch size limit, the Profiler employs the AAN-LL strategy to designate auxiliary networks to the layers of the input CNN. Subsequently, it benchmarks the GPU memory required for training each layer for various batch sizes. We observe that the GPU memory used by each layer linearly correlates with the batch size as shown in Figure~\ref{fig:linear_model}\footnote{A similar trend is noted for larger VGG and ResNet variants but we only provide results for VGG11 to clearly show the linear relationship.}. By leveraging this observation, the Profiler constructs layer-wise linear models to predict the GPU memory requirements for a given batch size (§1).

\textbf{Partitioner}:
This module segments the input CNN into distinct blocks by grouping layers based on their GPU memory consumption as shown in Algorithm~\ref{alg:partitioning} (§2). Initially, the Partitioner computes the maximum possible batch size each layer can accommodate while adhering to the GPU memory budget, leveraging the linear models produced by the Profiler (refer to lines 2-3). Although the later CNN layers may be capable of supporting batch sizes in the order of thousands as illustrated in Figure~\ref{fig:max_batch_size}, traditional CNN training often adopts batch sizes between 32 and 512~\cite{keskar2017on}.
Employing excessively large batch sizes during training might inadvertently impair generalization~\cite{masters2018revisiting, keskar2017on}. As a remedial strategy, the Partitioner caps the batch size at a predefined limit if its calculated value surpasses this threshold, as specified in line 4. Such a restriction on batch size not only guarantees the feasibility of on-device training but also preserves the ability of the model to generalize. 

The Partitioner subsequently utilizes the calculated maximum batch size for each layer as a basis for grouping layers. If the variation in feasible batch sizes between successive layers remains below a 40\% margin (defined as the grouping threshold $\rho$), then they are grouped into one block, as indicated in line 12. The batch size designated to this block defaults to the smallest batch size calculated for its constituent layers, as denoted in line 11. It was empirically found that the 40\% threshold is the most effective in balancing training efficiency and model convergence across thresholds spanning 10\% to 70\%. This choice of threshold is based on an exhaustive analysis, which balances the equilibrium between layer grouping and GPU memory optimization during training. Importantly, blocks may consist of a single layer if the batch sizes of its neighbouring layers are above the threshold.

\begin{algorithm}[t]
\caption{CNN Partitioning in \accordion{}}
\label{alg:partitioning} 
\begin{algorithmic}[1] 
\REQUIRE GPU memory budget $M$
\REQUIRE Batch size limit $B$
\REQUIRE Layer-wise linear regression models $R$
\REQUIRE Grouping threshold $\rho$

\ENSURE Partitioned CNN blocks and their respective batch sizes

\STATE Initialize empty list of blocks $Blocks$
\FOR{each layer $l$ in CNN}
    \STATE $t \gets \text{max batch size from } R[l] \text{ such that GPU memory} \leq M$
    \STATE $b_{l} \gets \min(t, B)$
\ENDFOR

\FOR{$i = 1$ to number of layers in CNN}
    \STATE Initialize current block $Block \gets \{ \}$
    \STATE $Block.\text{layers} \gets [i]$
    \STATE $Block.\text{batch\_size} \gets b_{i}$
    \WHILE{$i+1 \leq \text{number of layers}$ \AND $|b_{i+1} - b_{i}| \leq \rho \times b_{i}$}
        \STATE $Block.\text{batch\_size} \gets \min(Block.\text{batch\_size}, b_{i+1})$
        \STATE Append $i+1$ to $Block.\text{layers}$
        \STATE Increment $i$
    \ENDWHILE
    \STATE Append $Block$ to $Blocks$
\ENDFOR

\STATE \textbf{return} $Blocks$

\end{algorithmic}
\end{algorithm}

\textbf{Controller and Worker}:
Following the partitioning phase, the Controller designates the training task of the primary block to a specific worker (§3). Once allocated, the worker imports the specified block into its working GPU memory to start training. The other blocks are stored in the storage device at this time (§3.1).

Considering a training set, denoted by \( D \triangleq \{x^{k},y^{k}\}_{k<K} \) and with a batch size ascertained by the Partitioner, the worker fetches the training batches into its active GPU memory (§3.2). In tandem, layer-wise training for the layers encapsulated within the block is initiated, adhering to the protocol described in Algorithm~\ref{alg:block_local_learning}. The convergence analysis of adaptive local learning underpinning \accordion{} is presented in Appendix B.

During this training phase, the training batches are provided to the block as indicated in lines 1-2. The training batch is processed by the initial layer and produces the activations as specified in line 3. Subsequently, these activations are relayed to the auxiliary network, resulting in the creation of local predictions as shown in line 4. These predictions are used to compute the loss in line 5. The loss then determines the parameter gradients shown in line 6, which subsequently guide the required parameter updates shown in line 7. Following this update, the generated activations progress from the initial layer to the subsequent layer. This layer-specific procedure ensures that each layer not only processes the data but also updates its parameters. This iterative sequence is continued until the entire batch passes through every layer.

\begin{algorithm}[t]
\caption{Block-wise Local Learning in \accordion{}}
\label{alg:block_local_learning}
\begin{algorithmic}[1]
\REQUIRE Current $Block$ with \( N \) layers and batch size \( B \) (assigned by the partitioner)

\REQUIRE Training batches \( D \triangleq \{x^{k},y^{k}\}_{k\leq K} \), where each batch contains \( B \) samples / activations
\REQUIRE Loss function $\ell$

\FOR{batch \( (x^{k}, y^{k}) \) in \( D \)}
    \FOR{\( n = 1 \) to \( N \)}
        
        \STATE $x_{n+1} \leftarrow \alpha P_{n} \theta_{n} x_{n}$ \COMMENT{Using Equation~\ref{eq:1}}
        
        \STATE Compute local prediction \( z_{n+1} \leftarrow A_{n} x_{n+1} \) \COMMENT{Using Equation~\ref{eq:2}}
        \STATE Compute loss \( \mathcal{L}_{n} \triangleq \ell(z_{n+1}, y^{k}) \) 
        \STATE Compute \( \nabla_{(\theta_{n}, \beta_{n}, \gamma_{n})} \mathcal{L}_{n} \)
        \STATE \( \theta_{n}, \beta_{n}, \gamma_{n} \leftarrow \) Update based on \( \nabla_{(\theta_{n}, \beta_{n} , \gamma_{n})} \mathcal{L}_{n} \)
        
    \ENDFOR
\ENDFOR

\end{algorithmic}
\end{algorithm}

\subsection{Efficient Forward Propagation via Prefetching and Adaptive Batching}
In LL, forward propagation determines the data flow within the CNN architecture. Each individual block within this architecture inherently depends on the outputs generated by its immediate predecessor. Thus, a computational challenge arises: when a block is being trained and requires inputs, the default mechanism is to execute forward propagation on the preceding blocks, even if they have already been trained. This process of generating inputs for an active training block from its already-trained predecessors is not only redundant, but also incurs significant computational overhead, resulting in longer training durations.

\begin{figure}[t]
  \includegraphics[width=0.49\textwidth,]{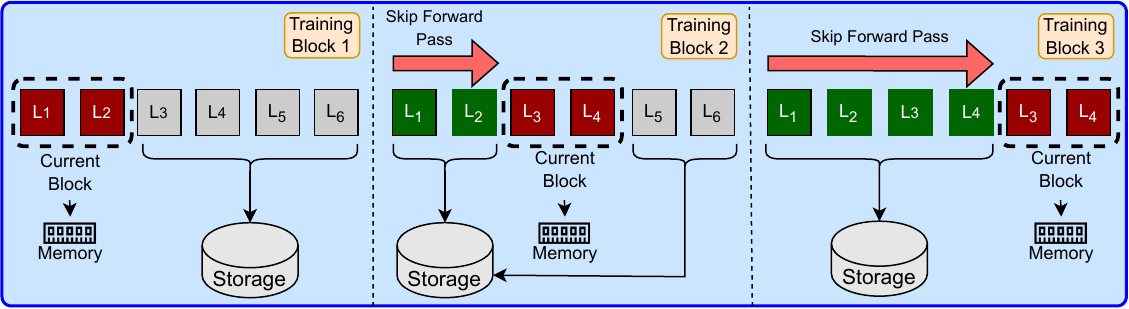}
\caption{Example of skipping forward pass and managing GPU memory in the \accordion{} training system for three time steps given a CNN with six layers. Red layers indicate the current block of layers being trained, green represents trained layers, and grey denotes untrained layers.}
  \label{fig:worker}
\end{figure}

\accordion{} addresses this challenge via a caching mechanism. When the training of a block is completed, the activations from the final layer of the block are transferred to a storage device (§3.3). Subsequently, the trained block is moved to storage and the next block is loaded into GPU memory (§3.1). The cached activations then become the input for the succeeding block. This design eliminates the need for redundant forward passes through already-trained blocks, thereby optimizing the training speed.
Figure \ref{fig:worker} shows GPU memory management during the training of blocks, illustrating where each block is stored and how the forward pass is skipped as training advances to deeper blocks.

Furthermore, \accordion{} integrates the AB-LL strategy, enhancing training performance. In this strategy (§3.2), as the activations are retrieved from storage in order to train a block, the number of activations in a given training batch is adjusted on-the-fly, making sure each block is trained with its designated batch size, allowing the current block to immediately use the training batch as input.

A key benefit of adaptive batch sizes is leveraging the unique GPU memory consumption patterns of each block. Initial blocks in a CNN tend to have higher GPU memory requirements, and thus, in edge computing environments with inherently limited working GPU memory, the batch sizes must be modest. However, as the training advances to subsequent blocks, which generally demand less GPU memory, the AB-LL strategy permits these blocks to handle larger batch sizes. 

AB-LL improves the computational efficiency by reducing the number of iterations required to process the entire dataset. By doing so, \textit{\accordion{} strikes a crucial balance: it significantly accelerates training while offering accuracy comparable to that of Backpropagation}.

\subsection{Trained Output CNN}
BP and classic LL produce a full-sized CNN after training that can be deployed for end-to-end inference. In contrast, \accordion{} evaluates the performance of each layer within all possible combinations of early exit models by considering validation accuracy and the number of parameters employed in a given early exit model. The outcome is a model that has an early exit point and achieves the highest validation accuracy while maintaining the smallest parameter count.

\begin{figure}[t]
    \centering
    \captionsetup{belowskip=-5pt} 
    \begin{tikzpicture}
        \begin{axis}[
            width=0.45\textwidth,
            height=0.19\textwidth, 
            ymin=0.2,
            ymax=1, 
            xlabel={Layer index},
            xtick={1,3,5,7,9,11,13},
            xticklabels={1,3,5,7,9,11,13},
            ylabel={Validation accuracy}, 
            font=\footnotesize,
            legend style={at={(0.5,-0.15)}, anchor=north, legend columns=-1},
            cycle list name=color list,
        ]
    
        \addplot[red, line width=1pt, mark=square*] coordinates {
            (1,0.46290)
            (2,0.51060)
            (3,0.60600)
            (4,0.64010)
            (5,0.65070)
            (6,0.64800)
            (7,0.64860)
            (8,0.64250)
            (9,0.63900)
            (10,0.63820)
            (11,0.63600)
            (12,0.63480)
            (13,0.63650)
        };

        \draw[dotted, line width=1pt] (5,0) -- (5,1);
        
        \draw[->, thick] (6.5,0.40)  -- (5.2,0.60) ;
        \node[right, font=\small] at (6.5,0.40) {Optimal Exit Point};
        \end{axis}
    \end{tikzpicture}
    \vspace{-10pt}
    \caption{Layer-wise validation accuracy of VGG-16 on CIFAR-100 in \accordion{}. Layer 5, the optimal exit point, achieves the highest validation accuracy with minimal parameters, making it the ideal early exit point.}
    \label{fig:overthinking}
    
\end{figure}
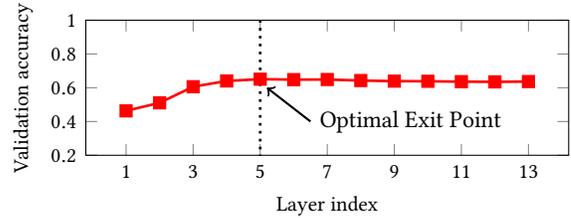

The output CNN model from \accordion{} is spatially efficient, as it contains only a fraction of the parameters compared to the original model. This is a consequence arising from the observation that the validation accuracy approaches a saturation point at a specific layer. Beyond this point, accuracy either remains consistent or decreases only trivially, as illustrated in Figure~\ref{fig:overthinking}. Such behavior resonates with findings from existing literature, captured under the term \textit{Overthinking}~\cite{SDN}. The concept highlights the diminishing returns of additional computational complexity in CNNs.

Conventionally, two approaches exist for early exit training. Each involves defining a few specific early exit points within a CNN. The first approach entails the simultaneous optimization of the entire CNN from its initialization, resulting in the computation of a weighted loss across all exit points~\cite{SDN}. In contrast, the second approach trains the base CNN until convergence. Then the early exit classifiers (auxiliary networks) undergo fine-tuning, while the parameters in the base CNN are frozen~\cite{HAPI}. Both approaches require at least the same amount of GPU memory as end-to-end BP.

\accordion{} detects overthinking at the layer level granularity, thereby mitigating the need for subsequent fine-tuning. In \accordion{}, every CNN layer is a prospective exit point, but the most suitable exit tailored to the distinct requirements of a given task is identified. Thus, the output CNN for on-device efficiency is optimized.

\section{Evaluation}
\label{sec:studies}
\begin{figure*}[t]
  \centering
  \includegraphics[width=0.95\textwidth]{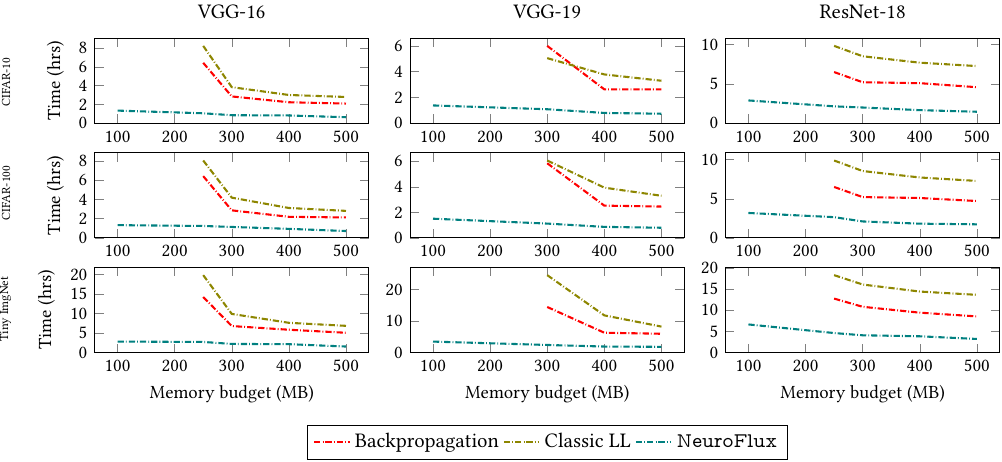}
    \vspace{-6pt}    
    \caption{Training time for different memory budgets using different training methods for VGG-16, VGG-19, and ResNet-18 models trained on the CIFAR-10, CIFAR-100, and the Tiny ImageNet datasets trained on the Nvidia AGX Orin.}
    \label{fig:training_time_comp}
\end{figure*}

\begin{figure*}[t]
  \centering
  \includegraphics[width=0.95\textwidth]{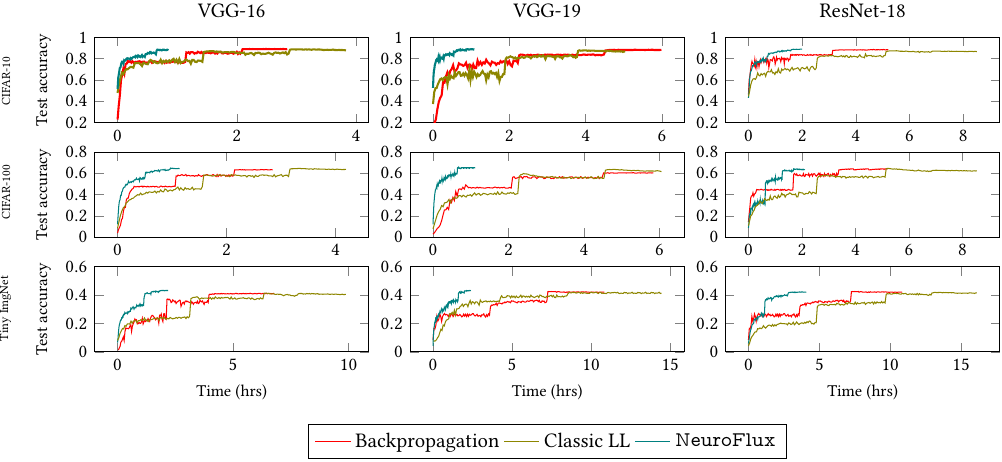}
    \vspace{-6pt}    
    \caption{Comparison of test accuracy as training proceeds for different training methods for VGG-16, VGG-19, and ResNet-18 models, trained on the CIFAR-10, CIFAR-100, and Tiny ImageNet datasets. Results are shown for a 300MB memory budget trained on the Nvidia AGX Orin.}
    \label{fig:training_time_acc}
\end{figure*}

\begin{table*}[h]
\centering
\small
\caption{Target hardware platforms}
\label{tab:platform-specs}
\begin{tabular}{|l|l|c|l|l|c|c|c|}
\hline
\multicolumn{1}{|c|}{\textbf{Platform}} &
  \multicolumn{1}{c|}{\textbf{CPU}} &
  \textbf{\# CPU Cores} &
  \multicolumn{1}{c|}{\textbf{Memory}} &
  \multicolumn{1}{c|}{\textbf{GPU}} &
  \textbf{\# GPU Cores} &
  \textbf{Peak TFLOPs} &
  \textbf{TDP} \\ \hline
Raspberry Pi 4B (Pi4B) & ARM Cortex-A72 & 4  & 4GB LPDDR4 & N/A & N/A & 0.00969 & 8W \\\hline
Nvidia Nano      & ARM Cortex-A57 & 4  & 4GB LPDDR4  & Maxwell & 128   & 0.472 & 5W      \\ \hline
Nvidia Xavier NX & ARM Carmel     & 6  & 8GB LPDDR4x & Volta   & 384   & 1.33  & 15W \\ \hline
Nvidia AGX ORIN  & ARM Carmel     & 12 & 64GB LPDDR5 & Ampere  & 1,536 & 4.76  & 50W   \\ \hline
\end{tabular}
\end{table*}

In this section, we evaluate the training performance of \accordion{} against BP and classic LL for varying GPU memory budgets and the inference performance across a range of edge devices. BP refers to vanilla Backpropagation, which includes no activation/gradient checkpointing \cite{gradcheckpointing}.

\subsection{Experimental Setup}
In our experimental setup, we specifically target three platforms, each characterized by distinct resource specifications, as presented in Table~\ref{tab:platform-specs}: the Nvidia Jetson Nano, Nvidia Jetson Xavier NX, and Nvidia Jetson AGX ORIN. Additionally, we employed the Raspberry Pi 4B to demonstrate the inference performance of the output model. To implement \accordion{}, we utilize PyTorch version 2.0.0 and torchvision version 0.2.0, with Python 3.11.

\textbf{Benchmarks}:
To assess the generalizability of our system, we conduct benchmark tests using a diverse set of CNNs that vary in depth and computational load. We include prominent VGG variants~\cite{VGG}, namely, VGG-16, and VGG-19, which are large and computationally intensive networks with conventional single-layer connectivity. We also employ ResNet-18~\cite{resnet}, from the residual network family, to demonstrate the feasibility of \accordion{}. We have selected large models to demonstrate the efficiency of our system under extreme memory and compute constraints.

\textbf{Datasets}:
\accordion{} is evaluated on three widely used image classification datasets: (1) CIFAR-10, (2) CIFAR-100 datasets~\cite{CIFAR10/100}, and (3) Tiny ImageNet dataset~\cite{tinyimagenet}. The datasets are used for image classification and offer a collection of images with varying degrees of complexity for evaluating the performance and accuracy of \accordion{}. The images of Tiny ImageNet were resized from 64$\times$64 pixels to 32$\times$32 pixels to match the input size of CIFAR datasets in order to consistently benchmark across the same CNNs. All measurements are averages of three independent experimental runs.

\subsection{End-to-End Training Performance}
In this section, we will present results to highlight the efficiency of \accordion{} in training a diverse range of CNNs under various GPU memory constraints. We consider both training time and the accuracy of \accordion{} relative to GPU memory budgets. Figure~\ref{fig:training_time_comp} shows the training time of BP, classic LL and \accordion{} for VGG-16, VGG-19 and ResNet-18 on the Nvidia Jetson AGX ORIN for GPU memory budgets ranging from 100MB to  500MB\footnote{Similar results are obtained on other hardware platforms but only show the results on one platform due to the limitation of space.}. 
\begin{tcolorbox}[
    width=0.48\textwidth,
    colframe=black!50!black,
    colback=gray!5]
\textbf{Observation 1:}
\accordion{} has a lower training time than BP for all GPU memory budgets.
\end{tcolorbox}

It is immediately noted from Figure~\ref{fig:training_time_comp} that \accordion{} achieves a reduction in training time - the speed-up ranges from 2.3$\times$ to 6.1$\times$ compared to BP and 3.3$\times$ to 10.3$\times$ compared to classic LL for the same GPU memory budgets.
Additionally, \accordion{} can successfully train CNNs under tight GPU memory constraints where BP and classic LL fail to do so. Therefore, there are no data points for BP and classic LL when the GPU memory budget is below 250MB for VGG-16 and ResNet-18 and below 300MB for VGG-19.
There is a performance gain when using \accordion{} for more generous GPU memory budgets, such as a 500MB limit.

\begin{tcolorbox}[
    width=0.48\textwidth,
    colframe=black!50!black,
    colback=gray!5]
\textbf{Observation 2:}
The performance of \accordion{} on lower GPU memory budgets is better than BP and classic LL on generous GPU memory budgets. It also successfully trains CNNs under GPU memory constraints that are unattainable by BP and classic LL.
\end{tcolorbox}

\accordion{} on a 100MB budget is 1.3$\times$ to 1.9$\times$ faster than BP and 2.1$\times$ to 2.5$\times$ faster than classic LL on a 500MB budget, hence reducing memory usage by 5$\times$. These results validate the efficiency of \accordion{} for CNN training in GPU memory-constrained environments. Figure~\ref{fig:training_time_acc} compares the test accuracy achieved in relation to the time taken for training using BP, classic LL, and \accordion. \accordion{} accelerates CNN training while the accuracy of the final model remains comparable to both BP and classic LL.

\begin{tcolorbox}[
    width=0.48\textwidth,
    colframe=black!50!black,
    colback=gray!5]
\textbf{Observation 3:}
For a given timeframe, \accordion{} achieves higher accuracy than BP and classic LL. 
\end{tcolorbox}

\begin{figure}
    \centering
    \captionsetup{belowskip=-5pt} 
    \begin{tikzpicture}
    \begin{groupplot}[
        group style={
            group size=2 by 1, 
            horizontal sep=1.4cm
        },
        width=0.27\textwidth,
        height=0.22\textwidth,
        ymin=0, ymax=17,
    legend style={at={(0.32,0.38)}, anchor=north, legend columns=1, font=\tiny, /tikz/every even column/.append style={column sep=0.5cm}, inner sep=0.15pt}
        ]

    \hspace{-12pt}
    \nextgroupplot[xlabel={Layer index}, xmin=1, xmax=17, xtick={1,3,...,17}, ylabel={Activation size}, ymode=log, ymin=1000, ymax=1000000, font=\small] 
    \addplot[very thin,color=red,mark=square*, mark size = 1pt] coordinates {
        (1,327680)
        (2,81920)
        (3,163840)
        (4,40960)
        (5,81920)
        (6,81920)
        (7,81920)
        (8,20480)
        (9,40960)
        (10,40960)
        (11,40960)
        (12,10240)
        (13,10240)
        (14,10240)
        (15,10240)
        (16,2560)
    };
    \addlegendentry{VGG-19}

    \addplot[very thin,color=blue,mark=triangle*, mark size = 1pt] coordinates {
        (1,327680)
        (2,327680)
        (3,327680)
        (4,327680)
        (5,327680)
        (6,163840)
        (7,163840)
        (8,163840)
        (9,163840)
        (10,81920)
        (11,81920)
        (12,81920)
        (13,81920)
        (14,40960)
        (15,40960)
        (16,40960)
        (17,2560)
    };
    \addlegendentry{ResNet-18}

    \nextgroupplot[xlabel={Layer index}, xmin=1, xmax=17, xtick={1,3,...,17}, ylabel style={align=center}, ylabel={Normalized FLOPs}, font=\small, ymode=linear, ymax=1.1]
    \addplot[very thin,color=red,mark=square*, mark size = 1pt] coordinates {
        (1,105.309/5292)
        (2,319.695/5292)
        (3,793.449/5292)
        (4,1077.885/5292)
        (5,1361.83/5292)
        (6,1740.147/5292)
        (7,2118.464/5292)
        (8,2355.121/5292)
        (9,2544.346/5292)
        (10,2827.943/5292)
        (11,3111.54/5292)
        (12,3324.353/5292)
        (13,3395.4442/5292)
        (14,3466.5354/5292)
        (15,3537.6266/5292)
        (16,3585.3757/5292)
        
    };

    \addplot[very thin,color=blue,mark=triangle*, mark size = 1pt] coordinates {
        (1,105.309/5292)
        (2,390.515/5292)
        (3,675.721/5292)
        (4,960.927/5292)
        (5,1246.133/5292)
        (6,1436.034/5292)
        (7,2004.159/5292)
        (8,2572.284/5292)
        (9,3140.409/5292)
        (10,3282.612/5292)
        (11,3660.929/5292)
        (12,4039.246/5292)
        (13,4417.563/5292)
        (14,4535.963/5292)
        (15,4819.56/5292)
        (16,5103.157/5292)
        (17,5292.617/5292)
    };

    \end{groupplot}
    \end{tikzpicture}
    \vspace{-24pt}
    \caption{Comparison of the activation size (total number of elements in the activation tensor) for different layers of VGG-19 and ResNet-18 (left). Normalized Cumulative FLOPs of the auxiliary networks for VGG-19 and ResNet-18 (right).}
    \label{fig:VGG_vs_RESNET}
\end{figure}
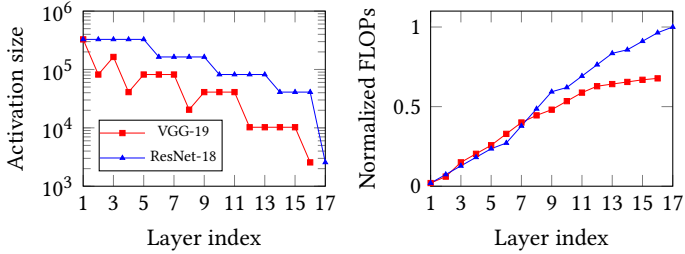

The inherent GPU memory efficiency of \accordion{} enables it to employ larger training batches compared to BP and classic LL, requiring fewer SGD steps across the entire training dataset. Thus, \accordion{} can reach its peak accuracy faster than the other methods. \accordion{} demonstrates better performance for VGG-19 over ResNet-18. VGG-19 frequently downsamples its activations, leading to smaller activation tensors, as shown in Figure~\ref{fig:VGG_vs_RESNET} (left). The computational requirements are influenced by the size of activations as they are processed by the auxiliary network during LL. This is highlighted in Figure~\ref{fig:VGG_vs_RESNET} (right) as the auxiliary networks of VGG-19 require fewer cumulative FLOPS compared to those of ResNet-18.

\subsection{Efficiency of Trained Output CNN}
Table~\ref{tab:early_exit_relative_compression} compares the size of the output model from BP, classic LL and \accordion{} when trained for different datasets. BP and classic LL produce CNNs of identical sizes. \accordion{} consistently yields highly efficient early exit CNNs that have 10.9$\times$ to 29.4$\times$ fewer parameters relative to the full-sized models generated by both BP and classic LL. Anecdotally, the compression levels attained by \accordion{} are similar to those achieved by structured pruning methods although the two are inherently different approaches~\cite{ProspectPF, DNNShifter, EfficientSP}.

\begin{tcolorbox}[
    width=0.48\textwidth,
    colframe=black!50!black,
    colback=gray!5]
\textbf{Observation 4:}
\accordion{} generates efficient CNNs that are spatially compressed and thus smaller in size than those obtained from BP and classic LL. 
\end{tcolorbox}

\begin{table}[h]
\centering
\caption{Parameter count of trained output CNNs.}
\footnotesize
\newcolumntype{B}{>{\columncolor[gray]{0.8}}c}
\begin{tabular}{|c|c|c|c|c|}
\hline
\multirow{2}{*}{\textbf{Dataset}} & \multirow{2}{*}{\textbf{Model}} & \multicolumn{2}{c|}{\textbf{\# Parameters (\(10^6 \))}} & \multirow{2}{*}{\vspace{10pt}\textbf{Compression}} \\
\cline{3-4}
 &  & \textbf{BP/LL} & \textbf{\accordion{}} &  \textbf{Factor}\\
\hline
\multirow{3}{*}{CIFAR-10} & VGG-16 & 14.7 & \textbf{0.5} & 29.4$\times$ \\
 & VGG-19 & 20.0 & \textbf{1.7} & 11.8$\times$ \\
 & ResNet-18 & 11.0 & \textbf{0.8} & 13.8$\times$ \\
\hline
\multirow{3}{*}{CIFAR-100} & VGG-16 & 14.7 & \textbf{1.19} & 12.3$\times$ \\
 & VGG-19 & 20.0 & \textbf{1.7} & 11.8$\times$ \\
 & ResNet-18 & 11.0 & \textbf{0.96} & 11.5$\times$ \\
\hline
\multirow{3}{*}{\vspace{15pt}Tiny} & VGG-16 & 14.7 & \textbf{1.19} & 12.3$\times$ \\
 ImageNet & VGG-19 & 20.0 & \textbf{1.19} & 16.8$\times$ \\
 & ResNet-18 & 11.0 & \textbf{1.01} & 10.9$\times$ \\
\hline

\end{tabular}
\label{tab:early_exit_relative_compression}
\end{table}

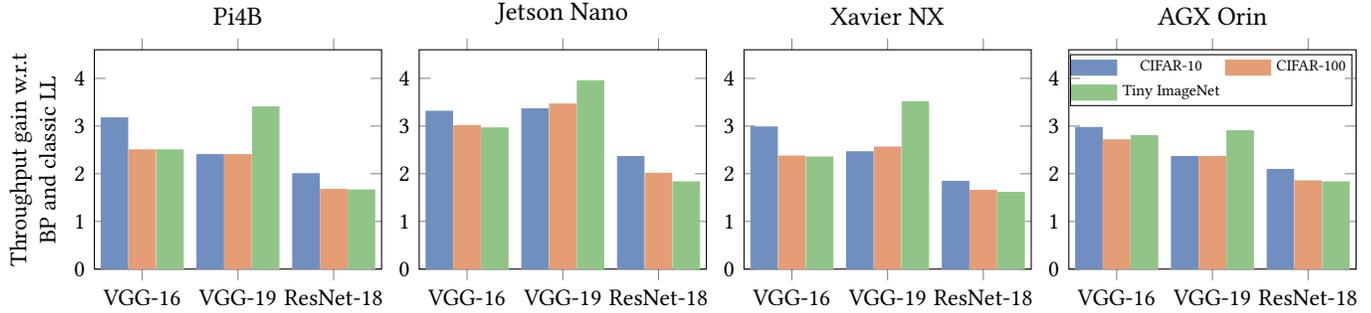
\begin{figure*}
\centering
\begin{tikzpicture}
\definecolor{color1}{RGB}{76,114,176} 
\definecolor{color2}{RGB}{221,132,82} 
\definecolor{color3}{RGB}{118,183,106} 

\begin{groupplot}[
    group style={
        group size=4 by 1,
        horizontal sep=0.5cm,
        group name=my plots
    },
    width=5.4cm,
    height=4.5cm, 
    ymin=0, 
    ymax=4, 
    ytick={0,1,2,3,4}, 
    ybar=0.5pt,
    enlarge x limits=0.25,
    enlarge y limits={0.15, upper},
    symbolic x coords={VGG-16, VGG-19, ResNet-18},
    xtick=data,
    label style={font=\small},
    tick label style={font=\small},
    cycle list={%
      {fill=color1!80, draw=color1!80},
      {fill=color2!80, draw=color2!80},
      {fill=color3!80, draw=color3!80}},
    legend columns=2,
legend style={
    at={(4.38,0.98)}, 
    anchor=north east,
    font=\tiny, inner sep=0.2pt}
]

\nextgroupplot[title=Pi4B, ylabel style={align=center}, ylabel= Throughput gain w.r.t\\BP and classic LL,legend entries={CIFAR-10, CIFAR-100, Tiny ImageNet}]
\addplot+[area legend] coordinates {
    (VGG-16, 3.17)
    (VGG-19, 2.40)
    (ResNet-18, 2.00)
};
\addplot+[area legend] coordinates {
    (VGG-16, 2.50)
    (VGG-19, 2.40)
    (ResNet-18, 1.67)
};
\addplot+[area legend] coordinates {
    (VGG-16, 2.5)
    (VGG-19, 3.4)
    (ResNet-18, 1.66)
};

\nextgroupplot[title=Jetson Nano]
\addplot coordinates {
    (VGG-16, 3.31)
    (VGG-19, 3.36)
    (ResNet-18, 2.36)
};
\addplot coordinates {
    (VGG-16, 3.01)
    (VGG-19, 3.46)
    (ResNet-18, 2.01)
};
\addplot+[area legend] coordinates {
    (VGG-16, 2.96)
    (VGG-19, 3.95)
    (ResNet-18, 1.83)
};

\nextgroupplot[title=Xavier NX]
\addplot coordinates {
    (VGG-16, 2.98)
    (VGG-19, 2.46)
    (ResNet-18, 1.84)
};
\addplot coordinates {
    (VGG-16, 2.37)
    (VGG-19, 2.56)
    (ResNet-18, 1.65)
};
\addplot+[area legend] coordinates {
    (VGG-16, 2.35)
    (VGG-19, 3.51)
    (ResNet-18, 1.61)
};

\nextgroupplot[title=AGX Orin]
\addplot coordinates {
    (VGG-16, 2.97)
    (VGG-19, 2.36)
    (ResNet-18, 2.09)
};
\addplot coordinates {
    (VGG-16, 2.71)
    (VGG-19, 2.36)
    (ResNet-18, 1.85)
};
\addplot+[area legend] coordinates {
    (VGG-16, 2.8)
    (VGG-19, 2.9)
    (ResNet-18, 1.83)
};

\end{groupplot}
\end{tikzpicture}
\vspace{-10pt}
\caption{Inference throughput gain achieved by \accordion{} over BP and classic LL for VGG-16, VGG-19 and ResNet-18 models on different hardware platforms using CIFAR-10, CIFAR-100 and Tiny ImageNet datasets.}
\label{fig:speedup_comparison}
\end{figure*}

This offers dual benefits: firstly, the parameter count of the model is smaller than BP and classic LL, and secondly, it provides an inference throughput gain. Thus, trained models from \accordion{} are suited for edge deployments, where computational and GPU memory resources are often limited. In Figure~\ref{fig:speedup_comparison}, we illustrate that there is a significant inference throughput gain achieved by the early exit model produced by \accordion{} when compared to the trained CNNs of BP and classic LL. The complete throughput data (images per second) is provided in Appendix~\ref{sec:appendix-inferencethroughput}.
It is immediately evident that the early exit model crafted by \accordion{} outperforms the conventional ones, delivering an increased throughput ranging from 1.61$\times$ to 3.95$\times$ on different target platforms.

\subsection{System Overheads}
The system overheads in \accordion{} arise from: 

\textbf{\textit{Profiler} and \textit{Partitioner} modules} construct layer-specific linear models to predict GPU memory consumption based on batch size. Subsequently, the CNN is partitioned into blocks. This processing overhead constitutes less than 1.5\% of the total CNN training time across all our experiments.

\textbf{\textit{Caching} and \textit{Prefetching} techniques}  utilize storage space to retain intermediate activations of trained blocks. These serve as inputs for training subsequent blocks, thereby eliminating the need for a forward pass over trained blocks. Datasets, such as CIFAR-10/100 and Tiny ImageNet, require approximately 0.2GB and 0.5GB of storage, respectively. Our caching strategy increases the storage need for activations to about 1.5$\times$ to 5.3$\times$ the size of the original dataset. Nonetheless, these storage demands are generally manageable in contemporary edge computing hardware, given their ample storage relative to working memory. Moreover, affordable storage mediums, such as high-speed SD cards, further ease any potential storage constraints.

\begin{tcolorbox}[
    width=0.48\textwidth,
    colframe=black!50!black,
    colback=gray!5]
\textbf{Observation 5:}
The performance gain from using \accordion{} substantially outweighs the overheads incurred by using its modules and underlying techniques.  
\end{tcolorbox}

\section{Related work}
\label{sec:related_work}

\textbf{Optimizing GPU memory usage during CNN training}: Three strategies considered below are used for optimizing GPU memory usage in BP-based training: 

\textit{Sparsity} - Sparse tensors are used in this method for representing model parameters to reduce GPU memory requirements, thereby compressing the model without significantly impacting accuracy~\cite{Hanprune, Guo_prune, Carreira_prune}. These methods either gradually train with sparsity~\cite{DynamicMP, Mostafa2019ParameterET} or start with inherently sparse networks~\cite{lth, synflow, DNNShifter}. However, sparse tensor operations can be slow due to the irregular distribution of non-zero elements, leading to inefficient data transfers between global GPU memory and computational registers~\cite{slow_sparse}.

\textit{Microbatching} - Smaller batch sizes are used in this method to reduce the GPU memory required for activations~\cite{GPipe, micro_batching}. This method is GPU memory efficient but requires precise hyperparameter tuning and increases the overall training time. This method may negatively impact the final accuracy of a model.

\textit{Gradient Checkpointing} - GPU memory is conserved in this method by saving only certain model activations. The trade-off is the recalculation of activations that are not saved, which results in longer training times~\cite{gradcheckpointing} than BP.

\textbf{Accelerating BP-based Training}: Two key approaches for enhancing the training speed and efficiency are considered next.

\textit{Parameter Freezing} - The concept of freezing parameters of the earlier layers of a CNN is known to improve training speed. For instance, SpotTune~\cite{SpotTune} employs pre-trained weights for CNN fine-tuning, and it determines which layers need to be trained on a per-sample basis. If a given sample does not require fine-tuning of the earlier CNN layers, then the parameter updates over those layers can be skipped, resulting in improved computational efficiency. However, the approach is limited to fine-tuning scenarios. In contrast, FreezeOut~\cite{freezeout} accelerates CNN training by systematically freezing layers and does not require backward passes over the frozen layers. This results in a 1.2$\times$ training speedup for ResNets, but shows no improvement for VGG models.

\textit{Caching Activations} - Egeria~\cite{Egeria} combines layer-freezing and the caching of intermediate results, thereby eliminating forward passes through the frozen layers. It improves the runtime by 1.19$\times$-1.43$\times$. However, Egeria is designed for HPC settings. To illustrate this, for facilitating training on GPUs Egeria requires a CPU-bound quantized reference model that identifies the layers to be frozen — a configuration unsuited for resource-constrained edge or mobile environments. Also, Egeria requires unfreezing layers when the learning rate reduces, as it relies on end-to-end BP. In contrast, \accordion{} underpinned by adaptive local learning ensures that the frozen layers are unaffected by learning rate changes of the subsequent layers. Lastly, PipeTransformer~\cite{PipeTransformer} is designed exclusively for Transformer models that utilize freezing progressive layers and caching intermediate results. The training pipeline is further optimized by omitting frozen layers, efficiently allocating active layers across fewer GPUs. This approach is suitable in HPC settings and is unattainable on edge or mobile devices due to inherent resource constraints, often limited to a single and small GPU. 

Building on the concept of optimizing computational efficiency, Skip-Convolutions~\cite{skipconv} re-engineer standard convolutions that are directly applied to residual frames. Using binary gates, the relevance of each layer to model prediction is evaluated — foreground elements are processed, while background elements may be bypassed to reduce computational overheads. Egeria~\cite{Egeria} explores the application of Skip-Convolutions as a heuristic during training to determine which layers to freeze, thereby reducing training time over Backpropagation by approximately 1.28$\times$. On the other hand, \accordion{} reduces training time by 2.3$\times$ to 6.1$\times$ under stringent GPU memory budgets.

\textbf{Comparing with Neural Architecture Search}: Neural Architecture Search (NAS) is aimed at automating the design of neural network architectures. Leading NAS methods such as DARTS~\cite{darts} and PreNAS~\cite{PreNAS} require extensive computational resources, typically ranging from 2 to 10 GPU days on server-grade hardware for CIFAR-10~\cite{DNNShifter}, which is a relatively small dataset, and produces models with an average parameter count ranging from 3.3M to 77M~\cite{DNNShifter}. In comparison, \textit{\accordion{}} produces a trained model in under two hours on the Nvidia Orin, an edge computing resource, with an average of 1M parameters for CIFAR-10. 

\textbf{Alternate Training Paradigms}: We focus on the below two paradigms that are developed as alternatives to BP-based training, which were introduced in Section~\ref{sec:background}.

\textit{Feedback Alignment} (FA): FA tackles the `weight transport problem' of DNN training~\cite{FA1, FA2}. Traditional BP-based training presumes a weight symmetry, where the backward pass employs the transposed weights from the forward pass. FA disrupts this symmetry: during the backward pass, it employs fixed random weights distinct from the forward pass. Although FA and its variants match BP in performance for smaller, fully connected networks, they do not work as well with CNNs~\cite{Signal_prop}.

\textit{Signal Propagation} (SP): SP employs a forward-pass training mechanism similar to the independent layer-wise approach used in LL, without the need for layer-specific auxiliary networks~\cite{Signal_prop}. SP utilizes a target generator that recasts target labels into input dimensions, thus generating the `context'. Each layer handles this context and the initial input, resulting in two discrete sets of activations. Predictions within layers are subsequently derived from the dot product of these activations, and the layer parameters are adjusted accordingly. While SP is over three times more GPU memory efficient than BP, it has lower training accuracy on small datasets, such as CIFAR-10. The inability to scale effectively to larger datasets, such as CIFAR-100 and Tiny ImageNet, limits its widespread adoption for training.

Both FA and SP have limitations, as mentioned in Section~\ref{sec:background} and highlighted in Figure~\ref{fig:comparison}. This positions LL, despite its GPU memory requirements, as the most viable alternative to traditional BP. \textit{\accordion{}} builds upon this foundation by developing adaptive local learning described in Section~\ref{sec:opportunities}. This approach not only addresses the GPU memory requirements inherent to LL but also significantly accelerates training compared to both classic LL and BP, while offering comparable accuracy across multiple datasets and CNN architectures, as empirically validated in Section~\ref{sec:studies}.

\section{Conclusions}
\label{sec:conclusions}
This paper demonstrates the feasibility of on-device CNN training using \accordion{}, both theoretically and experimentally. Its efficacy is not limited to vision-based tasks; \accordion{} also exhibits significant potential in training speech recognition models~\cite{REFL, Speech_1, Speech_2, Speech_3}. Another area is extending \accordion{} to support transformers, which are pivotal to NLP tasks. The complexity of training transformers is underscored by the GPU memory requirement (at least 8GB) for training BERT-small~\cite{bert}. 

Additionally, with the rise of federated learning for decentralized model training, \accordion{} offers a compelling advantage of improved training time. Efficient on-device training is essential to making federated learning feasible for edge and mobile devices. We envision that the training and memory efficiency offered by \accordion{} will enable the convergence of the global model more rapidly. This direction will extend the utility of \textit{\accordion{} in the context of distributed machine learning}.

While \accordion{} is designed for memory-constrained edge and mobile environments, it also offers benefits for desktop and server systems. These systems can also benefit from the reduced memory usage of \accordion{} and faster training times in scenarios when training large models on large datasets, such as ResNet-50~\cite{resnet} on ImageNet~\cite{ImageNet}. However, the extreme memory constraints that \accordion{} can operate in may not be of concern in larger desktop and server systems.

In summary, we introduced \accordion{}, a system optimized for on-device CNN training in memory-limited environments. Unlike traditional Backpropagation, \accordion{} uses adaptive local learning, showing consistent advancements across various CNNs and datasets. It reliably outperforms Backpropagation and classic local learning, with training speed improvements of 2.3$\times$ to 6.1$\times$  against Backpropagation and 3.3$\times$ to 10.3$\times$ against classic local learning for the same GPU memory budgets. \accordion{} generates a trained early exit model that reduces parameter counts by 10.9$\times$ to 29.4$\times$, boosting inference throughput by 1.61$\times$ to 3.95$\times$ on our experimental setup comprising diverse hardware platforms. \accordion{} provides a disruptive memory-efficient training solution that will enable the vision of practical on-device training in memory-constrained environments.

\section*{Acknowledgment}
We thank the anonymous reviewers and our shepherd Dr Hongyi Wang for their constructive feedback. Special thanks to Bailey Eccles, Di Wu, and Zihan Zhang from the Edge Computing Hub at the University of St Andrews for their critique and recommendations.

\bibliographystyle{ACM-Reference-Format}
\bibliography{EuroSys}


\begin{thebibliography}{76}


\ifx \showCODEN    \undefined \def \showCODEN     #1{\unskip}     \fi
\ifx \showDOI      \undefined \def \showDOI       #1{#1}\fi
\ifx \showISBNx    \undefined \def \showISBNx     #1{\unskip}     \fi
\ifx \showISBNxiii \undefined \def \showISBNxiii  #1{\unskip}     \fi
\ifx \showISSN     \undefined \def \showISSN      #1{\unskip}     \fi
\ifx \showLCCN     \undefined \def \showLCCN      #1{\unskip}     \fi
\ifx \shownote     \undefined \def \shownote      #1{#1}          \fi
\ifx \showarticletitle \undefined \def \showarticletitle #1{#1}   \fi
\ifx \showURL      \undefined \def \showURL       {\relax}        \fi
\providecommand\bibfield[2]{#2}
\providecommand\bibinfo[2]{#2}
\providecommand\natexlab[1]{#1}
\providecommand\showeprint[2][]{arXiv:#2}

\bibitem[Abbaschian et~al\mbox{.}(2021)]%
        {Speech_3}
\bibfield{author}{\bibinfo{person}{Babak~Joze Abbaschian}, \bibinfo{person}{Daniel Sierra-Sosa}, {and} \bibinfo{person}{Adel~Said Elmaghraby}.} \bibinfo{year}{2021}\natexlab{}.
\newblock \showarticletitle{Deep Learning Techniques for Speech Emotion Recognition, from Databases to Models}.
\newblock \bibinfo{journal}{\emph{Sensors}}.
\newblock


\bibitem[Abdelmoniem et~al\mbox{.}(2023)]%
        {REFL}
\bibfield{author}{\bibinfo{person}{Ahmed~M. Abdelmoniem}, \bibinfo{person}{Atal~Narayan Sahu}, \bibinfo{person}{Marco Canini}, {and} \bibinfo{person}{Suhaib~A. Fahmy}.} \bibinfo{year}{2023}\natexlab{}.
\newblock \showarticletitle{REFL: Resource-Efficient Federated Learning}. In \bibinfo{booktitle}{\emph{European Conference on Computer Systems}}.
\newblock


\bibitem[Akinpelu et~al\mbox{.}(2023)]%
        {Speech_2}
\bibfield{author}{\bibinfo{person}{Samson Akinpelu}, \bibinfo{person}{Serestina Viriri}, {and} \bibinfo{person}{Adekanmi Adegun}.} \bibinfo{year}{2023}\natexlab{}.
\newblock \showarticletitle{Lightweight Deep Learning Framework for Speech Emotion Recognition}.
\newblock \bibinfo{journal}{\emph{IEEE Access}}.
\newblock


\bibitem[Alizadeh et~al\mbox{.}(2022)]%
        {ProspectPF}
\bibfield{author}{\bibinfo{person}{Milad Alizadeh}, \bibinfo{person}{Shyam~A. Tailor}, \bibinfo{person}{Luisa~M Zintgraf}, \bibinfo{person}{Joost van Amersfoort}, \bibinfo{person}{Sebastian Farquhar}, \bibinfo{person}{Nicholas~Donald Lane}, {and} \bibinfo{person}{Yarin Gal}.} \bibinfo{year}{2022}\natexlab{}.
\newblock \showarticletitle{Prospect Pruning: Finding Trainable Weights at Initialization using Meta-Gradients}. In \bibinfo{booktitle}{\emph{International Conference on Machine Learning}}.
\newblock


\bibitem[Belilovsky et~al\mbox{.}(2019)]%
        {LL_Vanilla}
\bibfield{author}{\bibinfo{person}{Eugene Belilovsky}, \bibinfo{person}{Michael Eickenberg}, {and} \bibinfo{person}{Edouard Oyallon}.} \bibinfo{year}{2019}\natexlab{}.
\newblock \showarticletitle{{Greedy Layerwise Learning Can Scale To ImageNet}}. In \bibinfo{booktitle}{\emph{{International Conference on Machine Learning}}}.
\newblock


\bibitem[Belilovsky et~al\mbox{.}(2020)]%
        {DGL}
\bibfield{author}{\bibinfo{person}{Eugene Belilovsky}, \bibinfo{person}{Michael Eickenberg}, {and} \bibinfo{person}{Edouard Oyallon}.} \bibinfo{year}{2020}\natexlab{}.
\newblock \showarticletitle{Decoupled Greedy Learning of CNNs}. In \bibinfo{booktitle}{\emph{International Conference on Machine Learning}}.
\newblock


\bibitem[Bottou et~al\mbox{.}(2018)]%
        {Optimization_Methods}
\bibfield{author}{\bibinfo{person}{L\'{e}on Bottou}, \bibinfo{person}{Frank~E. Curtis}, {and} \bibinfo{person}{Jorge Nocedal}.} \bibinfo{year}{2018}\natexlab{}.
\newblock \showarticletitle{Optimization Methods for Large-Scale Machine Learning}.
\newblock \bibinfo{journal}{\emph{SIAM Rev.}}
\newblock


\bibitem[Brock et~al\mbox{.}(2017)]%
        {freezeout}
\bibfield{author}{\bibinfo{person}{Andrew Brock}, \bibinfo{person}{Theodore Lim}, \bibinfo{person}{J.~M. Ritchie}, {and} \bibinfo{person}{Nick Weston}.} \bibinfo{year}{2017}\natexlab{}.
\newblock \showarticletitle{FreezeOut: Accelerate Training by Progressively Freezing Layers}.
\newblock \bibinfo{journal}{\emph{arXiv:abs/1706.04983}}.
\newblock


\bibitem[Cai et~al\mbox{.}(2019)]%
        {proxylessnas}
\bibfield{author}{\bibinfo{person}{Han Cai}, \bibinfo{person}{Ligeng Zhu}, {and} \bibinfo{person}{Song Han}.} \bibinfo{year}{2019}\natexlab{}.
\newblock \showarticletitle{Proxyless{NAS}: Direct Neural Architecture Search on Target Task and Hardware}. In \bibinfo{booktitle}{\emph{International Conference on Learning Representations}}.
\newblock


\bibitem[Carreira-Perpinan and Idelbayev(2018)]%
        {Carreira_prune}
\bibfield{author}{\bibinfo{person}{Miguel~A. Carreira-Perpinan} {and} \bibinfo{person}{Yerlan Idelbayev}.} \bibinfo{year}{2018}\natexlab{}.
\newblock \showarticletitle{"Learning-Compression" Algorithms for Neural Net Pruning}. In \bibinfo{booktitle}{\emph{IEEE Conference on Computer Vision and Pattern Recognition}}.
\newblock


\bibitem[Chen et~al\mbox{.}(2016)]%
        {gradcheckpointing}
\bibfield{author}{\bibinfo{person}{Tianqi Chen}, \bibinfo{person}{Bing Xu}, \bibinfo{person}{Chiyuan Zhang}, {and} \bibinfo{person}{Carlos Guestrin}.} \bibinfo{year}{2016}\natexlab{}.
\newblock \showarticletitle{Training Deep Nets with Sublinear Memory Cost}.
\newblock \bibinfo{journal}{\emph{arXiv:abs/1604.06174}}.
\newblock


\bibitem[Deng et~al\mbox{.}(2009)]%
        {ImageNet}
\bibfield{author}{\bibinfo{person}{Jia Deng}, \bibinfo{person}{Wei Dong}, \bibinfo{person}{Richard Socher}, \bibinfo{person}{Li-Jia Li}, \bibinfo{person}{Kai Li}, {and} \bibinfo{person}{Li Fei-Fei}.} \bibinfo{year}{2009}\natexlab{}.
\newblock \showarticletitle{{ImageNet: A Large-Scale Hierarchical Image Database}}. In \bibinfo{booktitle}{\emph{IEEE Conference on Computer Vision and Pattern Recognition}}.
\newblock


\bibitem[Devlin et~al\mbox{.}(2019)]%
        {bert}
\bibfield{author}{\bibinfo{person}{Jacob Devlin}, \bibinfo{person}{Ming-Wei Chang}, \bibinfo{person}{Kenton Lee}, {and} \bibinfo{person}{Kristina Toutanova}.} \bibinfo{year}{2019}\natexlab{}.
\newblock \showarticletitle{{BERT}: Pre-training of Deep Bidirectional Transformers for Language Understanding}. In \bibinfo{booktitle}{\emph{Proceedings of the Conference of the North American Chapter of the Association for Computational Linguistics: Human Language Technologies}}.
\newblock


\bibitem[Eccles et~al\mbox{.}(2024)]%
        {DNNShifter}
\bibfield{author}{\bibinfo{person}{Bailey~J. Eccles}, \bibinfo{person}{Philip Rodgers}, \bibinfo{person}{Peter Kilpatrick}, \bibinfo{person}{Ivor Spence}, {and} \bibinfo{person}{Blesson Varghese}.} \bibinfo{year}{2024}\natexlab{}.
\newblock \showarticletitle{DNNShifter: An Efficient DNN Pruning System for Edge Computing}.
\newblock \bibinfo{journal}{\emph{Future Generation Computer Systems}}.
\newblock


\bibitem[Frankle and Carbin(2019)]%
        {lth}
\bibfield{author}{\bibinfo{person}{Jonathan Frankle} {and} \bibinfo{person}{Michael Carbin}.} \bibinfo{year}{2019}\natexlab{}.
\newblock \showarticletitle{The Lottery Ticket Hypothesis: Finding Sparse, Trainable Neural Networks}. In \bibinfo{booktitle}{\emph{International Conference on Learning Representations}}.
\newblock


\bibitem[Gim and Ko(2022)]%
        {SAGE}
\bibfield{author}{\bibinfo{person}{In Gim} {and} \bibinfo{person}{JeongGil Ko}.} \bibinfo{year}{2022}\natexlab{}.
\newblock \showarticletitle{Memory-efficient DNN training on mobile devices}. In \bibinfo{booktitle}{\emph{International Conference on Mobile Systems, Applications and Services}} \emph{(\bibinfo{series}{MobiSys '22})}.
\newblock


\bibitem[Guo et~al\mbox{.}(2021)]%
        {Guo_gesture}
\bibfield{author}{\bibinfo{person}{Junyao Guo}, \bibinfo{person}{Unmesh Kurup}, {and} \bibinfo{person}{Mohak Shah}.} \bibinfo{year}{2021}\natexlab{}.
\newblock \showarticletitle{Efficacy of Model Fine-Tuning for Personalized Dynamic Gesture Recognition}. In \bibinfo{booktitle}{\emph{Deep Learning for Human Activity Recognition}}.
\newblock


\bibitem[Guo et~al\mbox{.}(2019)]%
        {SpotTune}
\bibfield{author}{\bibinfo{person}{Yunhui Guo}, \bibinfo{person}{Honghui Shi}, \bibinfo{person}{Abhishek Kumar}, \bibinfo{person}{Kristen Grauman}, \bibinfo{person}{Tajana Rosing}, {and} \bibinfo{person}{Rogerio Feris}.} \bibinfo{year}{2019}\natexlab{}.
\newblock \showarticletitle{SpotTune: Transfer Learning Through Adaptive Fine-Tuning}. In \bibinfo{booktitle}{\emph{IEEE Conference on Computer Vision and Pattern Recognition}}.
\newblock


\bibitem[Guo et~al\mbox{.}(2016)]%
        {Guo_prune}
\bibfield{author}{\bibinfo{person}{Yiwen Guo}, \bibinfo{person}{Anbang Yao}, {and} \bibinfo{person}{Yurong Chen}.} \bibinfo{year}{2016}\natexlab{}.
\newblock \showarticletitle{Dynamic Network Surgery for Efficient DNNs}. In \bibinfo{booktitle}{\emph{International Conference on Neural Information Processing Systems}}.
\newblock


\bibitem[Gupta and Raskar(2018)]%
        {split2}
\bibfield{author}{\bibinfo{person}{Otkrist Gupta} {and} \bibinfo{person}{Ramesh Raskar}.} \bibinfo{year}{2018}\natexlab{}.
\newblock \showarticletitle{Distributed Learning of Deep Neural Network over Multiple Agents}.
\newblock \bibinfo{journal}{\emph{Journal of Network and Computer Applications}}.
\newblock


\bibitem[Habibian et~al\mbox{.}(2021)]%
        {skipconv}
\bibfield{author}{\bibinfo{person}{Amirhossein Habibian}, \bibinfo{person}{Davide Abati}, \bibinfo{person}{Taco Cohen}, {and} \bibinfo{person}{Babak~Ehteshami Bejnordi}.} \bibinfo{year}{2021}\natexlab{}.
\newblock \showarticletitle{Skip-Convolutions for Efficient Video Processing}. In \bibinfo{booktitle}{\emph{IEEE Conference on Computer Vision and Pattern Recognition}}.
\newblock


\bibitem[Han et~al\mbox{.}(2022)]%
        {splitgp}
\bibfield{author}{\bibinfo{person}{Dong-Jun Han}, \bibinfo{person}{Do-Yeon Kim}, \bibinfo{person}{Minseok Choi}, \bibinfo{person}{Christopher~G. Brinton}, {and} \bibinfo{person}{Jaekyun Moon}.} \bibinfo{year}{2022}\natexlab{}.
\newblock \showarticletitle{SplitGP: Achieving Both Generalization and Personalization in Federated Learning}.
\newblock \bibinfo{journal}{\emph{IEEE Conference on Computer Communications}}.
\newblock


\bibitem[Han et~al\mbox{.}(2016)]%
        {Hanprune}
\bibfield{author}{\bibinfo{person}{Song Han}, \bibinfo{person}{Huizi Mao}, {and} \bibinfo{person}{William~J. Dally}.} \bibinfo{year}{2016}\natexlab{}.
\newblock \showarticletitle{Deep Compression: Compressing Deep Neural Network with Pruning, Trained Quantization and Huffman Coding}. In \bibinfo{booktitle}{\emph{International Conference on Learning Representations}}.
\newblock


\bibitem[He et~al\mbox{.}(2021)]%
        {PipeTransformer}
\bibfield{author}{\bibinfo{person}{Chaoyang He}, \bibinfo{person}{Shen Li}, \bibinfo{person}{Mahdi Soltanolkotabi}, {and} \bibinfo{person}{Salman Avestimehr}.} \bibinfo{year}{2021}\natexlab{}.
\newblock \showarticletitle{PipeTransformer: Automated Elastic Pipelining for Distributed Training of Large-scale Models}. In \bibinfo{booktitle}{\emph{International Conference on Machine Learning}}.
\newblock


\bibitem[He et~al\mbox{.}(2020)]%
        {perception}
\bibfield{author}{\bibinfo{person}{Kaiming He}, \bibinfo{person}{Georgia Gkioxari}, \bibinfo{person}{Piotr Dollár}, {and} \bibinfo{person}{Ross Girshick}.} \bibinfo{year}{2020}\natexlab{}.
\newblock \showarticletitle{Mask R-CNN}.
\newblock \bibinfo{journal}{\emph{IEEE Transactions on Pattern Analysis and Machine Intelligence}}.
\newblock


\bibitem[He et~al\mbox{.}(2016)]%
        {resnet}
\bibfield{author}{\bibinfo{person}{Kaiming He}, \bibinfo{person}{Xiangyu Zhang}, \bibinfo{person}{Shaoqing Ren}, {and} \bibinfo{person}{Jian Sun}.} \bibinfo{year}{2016}\natexlab{}.
\newblock \showarticletitle{Deep Residual Learning for Image Recognition}. In \bibinfo{booktitle}{\emph{IEEE Conference on Computer Vision and Pattern Recognition}}.
\newblock


\bibitem[Howard et~al\mbox{.}(2017)]%
        {MobileNet}
\bibfield{author}{\bibinfo{person}{Andrew~G. Howard}, \bibinfo{person}{Menglong Zhu}, \bibinfo{person}{Bo Chen}, \bibinfo{person}{Dmitry Kalenichenko}, \bibinfo{person}{Weijun Wang}, \bibinfo{person}{Tobias Weyand}, \bibinfo{person}{Marco Andreetto}, {and} \bibinfo{person}{Hartwig Adam}.} \bibinfo{year}{2017}\natexlab{}.
\newblock \showarticletitle{MobileNets: Efficient Convolutional Neural Networks for Mobile Vision Applications}.
\newblock \bibinfo{journal}{\emph{arXiv:abs/1704.04861}}.
\newblock


\bibitem[Huang et~al\mbox{.}(2021)]%
        {real_world1}
\bibfield{author}{\bibinfo{person}{Baojin Huang}, \bibinfo{person}{Zhongyuan Wang}, \bibinfo{person}{Guangcheng Wang}, \bibinfo{person}{Kui Jiang}, \bibinfo{person}{Zheng He}, \bibinfo{person}{Hua Zou}, {and} \bibinfo{person}{Qin Zou}.} \bibinfo{year}{2021}\natexlab{}.
\newblock \showarticletitle{Masked Face Recognition Datasets and Validation}. In \bibinfo{booktitle}{\emph{2021 IEEE/CVF International Conference on Computer Vision Workshops}}.
\newblock


\bibitem[Huang et~al\mbox{.}(2018)]%
        {huang2018multiscale}
\bibfield{author}{\bibinfo{person}{Gao Huang}, \bibinfo{person}{Danlu Chen}, \bibinfo{person}{Tianhong Li}, \bibinfo{person}{Felix Wu}, \bibinfo{person}{Laurens van~der Maaten}, {and} \bibinfo{person}{Kilian Weinberger}.} \bibinfo{year}{2018}\natexlab{}.
\newblock \showarticletitle{Multi-Scale Dense Networks for Resource Efficient Image Classification}. In \bibinfo{booktitle}{\emph{International Conference on Learning Representations}}.
\newblock


\bibitem[Huang et~al\mbox{.}(2019)]%
        {GPipe}
\bibfield{author}{\bibinfo{person}{Yanping Huang}, \bibinfo{person}{Youlong Cheng}, \bibinfo{person}{Ankur Bapna}, \bibinfo{person}{Orhan Firat}, \bibinfo{person}{Mia~Xu Chen}, \bibinfo{person}{Dehao Chen}, \bibinfo{person}{HyoukJoong Lee}, \bibinfo{person}{Jiquan Ngiam}, \bibinfo{person}{Quoc~V. Le}, \bibinfo{person}{Yonghui Wu}, {and} \bibinfo{person}{Zhifeng Chen}.} \bibinfo{year}{2019}\natexlab{}.
\newblock \showarticletitle{GPipe: Efficient Training of Giant Neural Networks Using Pipeline Parallelism}.
\newblock \bibinfo{journal}{\emph{International Conference on Neural Information Processing Systems}}.
\newblock


\bibitem[Huynh et~al\mbox{.}(2021)]%
        {iMon}
\bibfield{author}{\bibinfo{person}{Sinh Huynh}, \bibinfo{person}{Rajesh Balan}, {and} \bibinfo{person}{Jeonggil Ko}.} \bibinfo{year}{2021}\natexlab{}.
\newblock \showarticletitle{iMon: Appearance-based Gaze Tracking System on Mobile Devices}. In \bibinfo{booktitle}{\emph{Proceedings of the ACM on Interactive, Mobile, Wearable and Ubiquitous Technologies}}.
\newblock


\bibitem[Ioffe(2017)]%
        {micro_batching}
\bibfield{author}{\bibinfo{person}{Sergey Ioffe}.} \bibinfo{year}{2017}\natexlab{}.
\newblock \showarticletitle{Batch Renormalization: Towards Reducing Minibatch Dependence in Batch-Normalized Models}. In \bibinfo{booktitle}{\emph{Advances in Neural Information Processing Systems}}.
\newblock


\bibitem[Iv et~al\mbox{.}(2017)]%
        {face_Luttrell2}
\bibfield{author}{\bibinfo{person}{Joseph Bailey~Luttrell Iv}, \bibinfo{person}{Zhaoxian Zhou}, \bibinfo{person}{Chaoyang Zhang}, \bibinfo{person}{Ping Gong}, {and} \bibinfo{person}{Yuanyuan Zhang}.} \bibinfo{year}{2017}\natexlab{}.
\newblock \showarticletitle{Facial Recognition via Transfer Learning: Fine-Tuning Keras\_vggface}.
\newblock \bibinfo{journal}{\emph{International Conference on Computational Science and Computational Intelligence}}.
\newblock


\bibitem[Kaya et~al\mbox{.}(2018)]%
        {SDN}
\bibfield{author}{\bibinfo{person}{Yigitcan Kaya}, \bibinfo{person}{Sanghyun Hong}, {and} \bibinfo{person}{Tudor Dumitras}.} \bibinfo{year}{2018}\natexlab{}.
\newblock \showarticletitle{Shallow-Deep Networks: Understanding and Mitigating Network Overthinking}. In \bibinfo{booktitle}{\emph{International Conference on Machine Learning}}.
\newblock


\bibitem[Kemelmacher-Shlizerman et~al\mbox{.}(2016)]%
        {real_world3}
\bibfield{author}{\bibinfo{person}{Ira Kemelmacher-Shlizerman}, \bibinfo{person}{Steven~M Seitz}, \bibinfo{person}{Daniel Miller}, {and} \bibinfo{person}{Evan Brossard}.} \bibinfo{year}{2016}\natexlab{}.
\newblock \showarticletitle{The MegaFace Benchmark: 1 Million Faces for Recognition at Scale}. In \bibinfo{booktitle}{\emph{IEEE Conference on Computer Vision and Pattern Recognition}}.
\newblock


\bibitem[Keskar et~al\mbox{.}(2017)]%
        {keskar2017on}
\bibfield{author}{\bibinfo{person}{Nitish~Shirish Keskar}, \bibinfo{person}{Dheevatsa Mudigere}, \bibinfo{person}{Jorge Nocedal}, \bibinfo{person}{Mikhail Smelyanskiy}, {and} \bibinfo{person}{Ping Tak~Peter Tang}.} \bibinfo{year}{2017}\natexlab{}.
\newblock \showarticletitle{On Large-Batch Training for Deep Learning: Generalization Gap and Sharp Minima}. In \bibinfo{booktitle}{\emph{International Conference on Learning Representations}}.
\newblock


\bibitem[Kohan et~al\mbox{.}(2023)]%
        {Signal_prop}
\bibfield{author}{\bibinfo{person}{Adam Kohan}, \bibinfo{person}{Edward~A. Rietman}, {and} \bibinfo{person}{Hava~T. Siegelmann}.} \bibinfo{year}{2023}\natexlab{}.
\newblock \showarticletitle{Signal Propagation: The Framework for Learning and Inference in a Forward Pass}.
\newblock \bibinfo{journal}{\emph{IEEE Transactions on Neural Networks and Learning Systems}}.
\newblock


\bibitem[Kouris and Bouganis(2018)]%
        {decision_making}
\bibfield{author}{\bibinfo{person}{Alexandros Kouris} {and} \bibinfo{person}{Christos-Savvas Bouganis}.} \bibinfo{year}{2018}\natexlab{}.
\newblock \showarticletitle{Learning to Fly by MySelf: A Self-Supervised CNN-Based Approach for Autonomous Navigation}. In \bibinfo{booktitle}{\emph{IEEE/RSJ International Conference on Intelligent Robots and Systems}}.
\newblock


\bibitem[Krizhevsky(2009)]%
        {CIFAR10/100}
\bibfield{author}{\bibinfo{person}{Alex Krizhevsky}.} \bibinfo{year}{2009}\natexlab{}.
\newblock \showarticletitle{Learning Multiple Layers of Features from Tiny Images}.
\newblock \bibinfo{journal}{\emph{Technical Report}}.
\newblock
\newblock
\shownote{https://www.cs.toronto.edu/~kriz/cifar.html}.


\bibitem[Laskaridis et~al\mbox{.}(2020)]%
        {HAPI}
\bibfield{author}{\bibinfo{person}{Stefanos Laskaridis}, \bibinfo{person}{Stylianos~I. Venieris}, \bibinfo{person}{Hyeji Kim}, {and} \bibinfo{person}{Nicholas~D. Lane}.} \bibinfo{year}{2020}\natexlab{}.
\newblock \showarticletitle{HAPI: Hardware-Aware Progressive Inference}. In \bibinfo{booktitle}{\emph{International Conference on Computer-Aided Design}}.
\newblock


\bibitem[Le and Yang(2015)]%
        {tinyimagenet}
\bibfield{author}{\bibinfo{person}{Ya Le} {and} \bibinfo{person}{Xuan~S. Yang}.} \bibinfo{year}{2015}\natexlab{}.
\newblock \showarticletitle{Tiny ImageNet Visual Recognition Challenge}.
\newblock
\urldef\tempurl%
\url{http://vision.stanford.edu/teaching/cs231n/reports/2015/pdfs/yle_project.pdf}
\showURL{%
\tempurl}


\bibitem[Lee et~al\mbox{.}(2019)]%
        {mobilegpu}
\bibfield{author}{\bibinfo{person}{Juhyun Lee}, \bibinfo{person}{Nikolay Chirkov}, \bibinfo{person}{Ekaterina Ignasheva}, \bibinfo{person}{Yury Pisarchyk}, \bibinfo{person}{Mogan Shieh}, \bibinfo{person}{Fabio Riccardi}, \bibinfo{person}{Raman Sarokin}, \bibinfo{person}{Andrei Kulik}, {and} \bibinfo{person}{Matthias Grundmann}.} \bibinfo{year}{2019}\natexlab{}.
\newblock \showarticletitle{On-Device Neural Net Inference with Mobile GPUs}.
\newblock \bibinfo{journal}{\emph{arXiv:abs/1907.01989}}.
\newblock


\bibitem[Li et~al\mbox{.}(2019)]%
        {9010043}
\bibfield{author}{\bibinfo{person}{H. Li}, \bibinfo{person}{H. Zhang}, \bibinfo{person}{X. Qi}, \bibinfo{person}{Y. Ruigang}, {and} \bibinfo{person}{G. Huang}.} \bibinfo{year}{2019}\natexlab{}.
\newblock \showarticletitle{Improved Techniques for Training Adaptive Deep Networks}. In \bibinfo{booktitle}{\emph{IEEE/CVF International Conference on Computer Vision}}.
\newblock


\bibitem[Liao et~al\mbox{.}(2016)]%
        {FA2}
\bibfield{author}{\bibinfo{person}{Qianli Liao}, \bibinfo{person}{Joel~Z. Leibo}, {and} \bibinfo{person}{Tomaso Poggio}.} \bibinfo{year}{2016}\natexlab{}.
\newblock \showarticletitle{How Important is Weight Symmetry in Backpropagation?}. In \bibinfo{booktitle}{\emph{AAAI Conference on Artificial Intelligence}}.
\newblock


\bibitem[Lillicrap et~al\mbox{.}(2016)]%
        {FA1}
\bibfield{author}{\bibinfo{person}{Timothy~P. Lillicrap}, \bibinfo{person}{Daniel Cownden}, \bibinfo{person}{Douglas~Blair Tweed}, {and} \bibinfo{person}{Colin~J. Akerman}.} \bibinfo{year}{2016}\natexlab{}.
\newblock \showarticletitle{Random Synaptic Feedback Weights Support Error Backpropagation for Deep Learning}.
\newblock \bibinfo{journal}{\emph{Nature Communications}}.
\newblock


\bibitem[Lin et~al\mbox{.}(2020)]%
        {DynamicMP}
\bibfield{author}{\bibinfo{person}{Tao Lin}, \bibinfo{person}{Sebastian~U. Stich}, \bibinfo{person}{Luis Barba}, \bibinfo{person}{Daniil Dmitriev}, {and} \bibinfo{person}{Martin Jaggi}.} \bibinfo{year}{2020}\natexlab{}.
\newblock \showarticletitle{Dynamic Model Pruning with Feedback}. In \bibinfo{booktitle}{\emph{International Conference on Learning Representations}}.
\newblock


\bibitem[Liu et~al\mbox{.}(2019)]%
        {darts}
\bibfield{author}{\bibinfo{person}{Hanxiao Liu}, \bibinfo{person}{Karen Simonyan}, {and} \bibinfo{person}{Yiming Yang}.} \bibinfo{year}{2019}\natexlab{}.
\newblock \showarticletitle{{DARTS}: Differentiable Architecture Search}. In \bibinfo{booktitle}{\emph{International Conference on Learning Representations}}.
\newblock


\bibitem[Luttrell et~al\mbox{.}(2018)]%
        {face_Luttrell}
\bibfield{author}{\bibinfo{person}{Joseph Luttrell}, \bibinfo{person}{Zhaoxian Zhou}, \bibinfo{person}{Yuanyuan Zhang}, \bibinfo{person}{Chaoyang Zhang}, \bibinfo{person}{Ping Gong}, \bibinfo{person}{Bei Yang}, {and} \bibinfo{person}{Runzhi Li}.} \bibinfo{year}{2018}\natexlab{}.
\newblock \showarticletitle{A Deep Transfer Learning Approach to Fine-Tuning Facial Recognition Models}. In \bibinfo{booktitle}{\emph{IEEE Conference on Industrial Electronics and Applications}}.
\newblock


\bibitem[Mandal et~al\mbox{.}(2021)]%
        {real_world2}
\bibfield{author}{\bibinfo{person}{Bishwas Mandal}, \bibinfo{person}{Adaeze Okeukwu}, {and} \bibinfo{person}{Yihong Theis}.} \bibinfo{year}{2021}\natexlab{}.
\newblock \showarticletitle{Masked Face Recognition using ResNet-50}.
\newblock \bibinfo{journal}{\emph{arXiv:abs/2104.08997}}.
\newblock


\bibitem[Masters and Luschi(2018)]%
        {masters2018revisiting}
\bibfield{author}{\bibinfo{person}{Dominic Masters} {and} \bibinfo{person}{Carlo Luschi}.} \bibinfo{year}{2018}\natexlab{}.
\newblock \showarticletitle{Revisiting Small Batch Training for Deep Neural Networks}.
\newblock \bibinfo{journal}{\emph{arXiv:abs/1804.07612}}.
\newblock


\bibitem[McMahan et~al\mbox{.}(2017)]%
        {com1}
\bibfield{author}{\bibinfo{person}{Brendan McMahan}, \bibinfo{person}{Eider Moore}, \bibinfo{person}{Daniel Ramage}, \bibinfo{person}{Seth Hampson}, {and} \bibinfo{person}{Blaise Aguera~y Arcas}.} \bibinfo{year}{2017}\natexlab{}.
\newblock \showarticletitle{{Communication-Efficient Learning of Deep Networks from Decentralized Data}}. In \bibinfo{booktitle}{\emph{International Conference on Artificial Intelligence and Statistics}}.
\newblock


\bibitem[Molchanov et~al\mbox{.}(2017)]%
        {molchanov2017pruning}
\bibfield{author}{\bibinfo{person}{Pavlo Molchanov}, \bibinfo{person}{Stephen Tyree}, \bibinfo{person}{Tero Karras}, \bibinfo{person}{Timo Aila}, {and} \bibinfo{person}{Jan Kautz}.} \bibinfo{year}{2017}\natexlab{}.
\newblock \showarticletitle{Pruning Convolutional Neural Networks for Resource Efficient Inference}. In \bibinfo{booktitle}{\emph{International Conference on Learning Representations}}.
\newblock


\bibitem[Mostafa and Wang(2019)]%
        {Mostafa2019ParameterET}
\bibfield{author}{\bibinfo{person}{Hesham Mostafa} {and} \bibinfo{person}{Xin Wang}.} \bibinfo{year}{2019}\natexlab{}.
\newblock \showarticletitle{Parameter Efficient Training of Deep Convolutional Neural Networks by Dynamic Sparse Reparameterization}. In \bibinfo{booktitle}{\emph{International Conference on Machine Learning}}.
\newblock


\bibitem[Padi et~al\mbox{.}(2021)]%
        {Speech_1}
\bibfield{author}{\bibinfo{person}{Sarala Padi}, \bibinfo{person}{Seyed~Omid Sadjadi}, \bibinfo{person}{Dinesh Manocha}, {and} \bibinfo{person}{Ram~D. Sriram}.} \bibinfo{year}{2021}\natexlab{}.
\newblock \showarticletitle{Improved Speech Emotion Recognition using Transfer Learning and Spectrogram Augmentation}.
\newblock \bibinfo{journal}{\emph{Proceedings of the International Conference on Multimodal Interaction}}.
\newblock


\bibitem[Parisi et~al\mbox{.}(2019)]%
        {PARISI201954}
\bibfield{author}{\bibinfo{person}{German~I. Parisi}, \bibinfo{person}{Ronald Kemker}, \bibinfo{person}{Jose~L. Part}, \bibinfo{person}{Christopher Kanan}, {and} \bibinfo{person}{Stefan Wermter}.} \bibinfo{year}{2019}\natexlab{}.
\newblock \showarticletitle{Continual Lifelong Learning with Neural Networks: A Review}.
\newblock \bibinfo{journal}{\emph{Neural Networks}}.
\newblock


\bibitem[Park et~al\mbox{.}(2021)]%
        {Sign_Language}
\bibfield{author}{\bibinfo{person}{HyeonJung Park}, \bibinfo{person}{Youngki Lee}, {and} \bibinfo{person}{JeongGil Ko}.} \bibinfo{year}{2021}\natexlab{}.
\newblock \showarticletitle{Enabling Real-time Sign Language Translation on Mobile Platforms with On-board Depth Cameras}. In \bibinfo{booktitle}{\emph{Proceedings of the ACM on Interactive, Mobile, Wearable and Ubiquitous Technologies.}}
\newblock


\bibitem[Patterson et~al\mbox{.}(2022)]%
        {Carbon_Footprint}
\bibfield{author}{\bibinfo{person}{David Patterson}, \bibinfo{person}{Joseph Gonzalez}, \bibinfo{person}{Urs Hölzle}, \bibinfo{person}{Quoc Le}, \bibinfo{person}{Chen Liang}, \bibinfo{person}{Lluis-Miquel Munguia}, \bibinfo{person}{Daniel Rothchild}, \bibinfo{person}{David~R. So}, \bibinfo{person}{Maud Texier}, {and} \bibinfo{person}{Jeff Dean}.} \bibinfo{year}{2022}\natexlab{}.
\newblock \showarticletitle{The Carbon Footprint of Machine Learning Training Will Plateau, Then Shrink}.
\newblock \bibinfo{journal}{\emph{Computer}} (\bibinfo{year}{2022}).
\newblock


\bibitem[Pham et~al\mbox{.}(2018)]%
        {NAS2}
\bibfield{author}{\bibinfo{person}{Hieu Pham}, \bibinfo{person}{Melody Guan}, \bibinfo{person}{Barret Zoph}, \bibinfo{person}{Quoc Le}, {and} \bibinfo{person}{Jeff Dean}.} \bibinfo{year}{2018}\natexlab{}.
\newblock \showarticletitle{Efficient Neural Architecture Search via Parameters Sharing}. In \bibinfo{booktitle}{\emph{International Conference on Machine Learning}}.
\newblock


\bibitem[Sahoo et~al\mbox{.}(2022)]%
        {s22030706}
\bibfield{author}{\bibinfo{person}{Jaya~Prakash Sahoo}, \bibinfo{person}{Allam~Jaya Prakash}, \bibinfo{person}{Paweł Pławiak}, {and} \bibinfo{person}{Saunak Samantray}.} \bibinfo{year}{2022}\natexlab{}.
\newblock \showarticletitle{Real-Time Hand Gesture Recognition Using Fine-Tuned Convolutional Neural Network}.
\newblock \bibinfo{journal}{\emph{Sensors}}.
\newblock


\bibitem[Sarfraz et~al\mbox{.}(2021)]%
        {Knowledge_Distillation}
\bibfield{author}{\bibinfo{person}{F. Sarfraz}, \bibinfo{person}{E. Arani}, {and} \bibinfo{person}{B. Zonooz}.} \bibinfo{year}{2021}\natexlab{}.
\newblock \showarticletitle{Knowledge Distillation Beyond Model Compression}. In \bibinfo{booktitle}{\emph{International Conference on Pattern Recognition}}.
\newblock


\bibitem[Schwartz et~al\mbox{.}(2020)]%
        {Green_AI}
\bibfield{author}{\bibinfo{person}{Roy Schwartz}, \bibinfo{person}{Jesse Dodge}, \bibinfo{person}{Noah~A. Smith}, {and} \bibinfo{person}{Oren Etzioni}.} \bibinfo{year}{2020}\natexlab{}.
\newblock \showarticletitle{Green AI}.
\newblock \bibinfo{journal}{\emph{Commun. ACM}} (\bibinfo{year}{2020}).
\newblock


\bibitem[Shi et~al\mbox{.}(2020)]%
        {slow_sparse}
\bibfield{author}{\bibinfo{person}{Shaohuai Shi}, \bibinfo{person}{Qiang Wang}, {and} \bibinfo{person}{Xiaowen Chu}.} \bibinfo{year}{2020}\natexlab{}.
\newblock \showarticletitle{Efficient Sparse-Dense Matrix-Matrix Multiplication on GPUs Using the Customized Sparse Storage Format}. In \bibinfo{booktitle}{\emph{IEEE International Conference on Parallel and Distributed Systems}}.
\newblock


\bibitem[Simonyan and Zisserman(2015)]%
        {VGG}
\bibfield{author}{\bibinfo{person}{Karen Simonyan} {and} \bibinfo{person}{Andrew Zisserman}.} \bibinfo{year}{2015}\natexlab{}.
\newblock \showarticletitle{Very Deep Convolutional Networks for Large-Scale Image Recognition}. In \bibinfo{booktitle}{\emph{International Conference on Learning Representations}}.
\newblock


\bibitem[Tanaka et~al\mbox{.}(2020)]%
        {synflow}
\bibfield{author}{\bibinfo{person}{Hidenori Tanaka}, \bibinfo{person}{Daniel Kunin}, \bibinfo{person}{Daniel~L Yamins}, {and} \bibinfo{person}{Surya Ganguli}.} \bibinfo{year}{2020}\natexlab{}.
\newblock \showarticletitle{Pruning neural networks without any data by iteratively conserving synaptic flow}. In \bibinfo{booktitle}{\emph{Advances in Neural Information Processing Systems}}.
\newblock


\bibitem[Teerapittayanon et~al\mbox{.}(2016)]%
        {BranchyNet}
\bibfield{author}{\bibinfo{person}{Surat Teerapittayanon}, \bibinfo{person}{Bradley McDanel}, {and} \bibinfo{person}{H.T. Kung}.} \bibinfo{year}{2016}\natexlab{}.
\newblock \showarticletitle{{BranchyNet: Fast Inference via Early Exiting from Deep Neural Networks}}. In \bibinfo{booktitle}{\emph{International Conference on Pattern Recognition}}.
\newblock


\bibitem[Vepakomma et~al\mbox{.}(2018)]%
        {split1}
\bibfield{author}{\bibinfo{person}{Praneeth Vepakomma}, \bibinfo{person}{Otkrist Gupta}, \bibinfo{person}{Tristan Swedish}, {and} \bibinfo{person}{Ramesh Raskar}.} \bibinfo{year}{2018}\natexlab{}.
\newblock \showarticletitle{Split Learning for Health: Distributed Deep Learning Without Sharing raw patient data}.
\newblock \bibinfo{journal}{\emph{arXiv:abs/1812.00564}}.
\newblock


\bibitem[Wang et~al\mbox{.}(2023a)]%
        {PreNAS}
\bibfield{author}{\bibinfo{person}{Haibin Wang}, \bibinfo{person}{Ce Ge}, \bibinfo{person}{Hesen Chen}, {and} \bibinfo{person}{Xiuyu Sun}.} \bibinfo{year}{2023}\natexlab{a}.
\newblock \showarticletitle{{PreNAS: Preferred One-Shot Learning Towards Efficient Neural Architecture Search}}. In \bibinfo{booktitle}{\emph{International Conference on Machine Learning}}.
\newblock


\bibitem[Wang et~al\mbox{.}(2020)]%
        {mobilegpu2}
\bibfield{author}{\bibinfo{person}{Siqi Wang}, \bibinfo{person}{Anuj Pathania}, {and} \bibinfo{person}{Tulika Mitra}.} \bibinfo{year}{2020}\natexlab{}.
\newblock \showarticletitle{Neural Network Inference on Mobile SoCs}.
\newblock \bibinfo{journal}{\emph{IEEE Design \& Test}}.
\newblock


\bibitem[Wang et~al\mbox{.}(2022)]%
        {Egeria}
\bibfield{author}{\bibinfo{person}{Yiding Wang}, \bibinfo{person}{Decang Sun}, \bibinfo{person}{Kai Chen}, \bibinfo{person}{Fan Lai}, {and} \bibinfo{person}{Mosharaf Chowdhury}.} \bibinfo{year}{2022}\natexlab{}.
\newblock \showarticletitle{Egeria: Efficient DNN Training with Knowledge-Guided Layer Freezing}.
\newblock \bibinfo{journal}{\emph{European Conference on Computer Systems}}.
\newblock


\bibitem[Wang et~al\mbox{.}(2023b)]%
        {split4}
\bibfield{author}{\bibinfo{person}{Zhiyuan Wang}, \bibinfo{person}{Hongli Xu}, \bibinfo{person}{Yang Xu}, \bibinfo{person}{Zhida Jiang}, {and} \bibinfo{person}{Jianchun Liu}.} \bibinfo{year}{2023}\natexlab{b}.
\newblock \showarticletitle{CoopFL: Accelerating Federated Learning with DNN Partitioning and Offloading in Heterogeneous Edge Computing}.
\newblock \bibinfo{journal}{\emph{Comput. Netw.}}
\newblock


\bibitem[Wu et~al\mbox{.}(2017)]%
        {Wu_2017_CVPR_Workshops}
\bibfield{author}{\bibinfo{person}{Bichen Wu}, \bibinfo{person}{Forrest Iandola}, \bibinfo{person}{Peter~H. Jin}, {and} \bibinfo{person}{Kurt Keutzer}.} \bibinfo{year}{2017}\natexlab{}.
\newblock \showarticletitle{SqueezeDet: Unified, Small, Low Power Fully Convolutional Neural Networks for Real-Time Object Detection for Autonomous Driving}. In \bibinfo{booktitle}{\emph{Proceedings of the IEEE Conference on Computer Vision and Pattern Recognition Workshops}}.
\newblock


\bibitem[Wu et~al\mbox{.}(2022)]%
        {fedadapt}
\bibfield{author}{\bibinfo{person}{Di Wu}, \bibinfo{person}{Rehmat Ullah}, \bibinfo{person}{Paul Harvey}, \bibinfo{person}{Peter Kilpatrick}, \bibinfo{person}{Ivor Spence}, {and} \bibinfo{person}{Blesson Varghese}.} \bibinfo{year}{2022}\natexlab{}.
\newblock \showarticletitle{FedAdapt: Adaptive Offloading for IoT Devices in Federated Learning}.
\newblock \bibinfo{journal}{\emph{IEEE Internet of Things Journal}}.
\newblock


\bibitem[Wu et~al\mbox{.}(2016)]%
        {quant1}
\bibfield{author}{\bibinfo{person}{Jiaxiang Wu}, \bibinfo{person}{Cong Leng}, \bibinfo{person}{Yuhang Wang}, \bibinfo{person}{Qinghao Hu}, {and} \bibinfo{person}{Jian Cheng}.} \bibinfo{year}{2016}\natexlab{}.
\newblock \showarticletitle{Quantized Convolutional Neural Networks for Mobile Devices}. In \bibinfo{booktitle}{\emph{IEEE Conference on Computer Vision and Pattern Recognition}}.
\newblock


\bibitem[Yu et~al\mbox{.}(2021)]%
        {EasiEdge}
\bibfield{author}{\bibinfo{person}{Fang Yu}, \bibinfo{person}{Li Cui}, \bibinfo{person}{Pengcheng Wang}, \bibinfo{person}{Chuanqi Han}, \bibinfo{person}{Ruoran Huang}, {and} \bibinfo{person}{Xi Huang}.} \bibinfo{year}{2021}\natexlab{}.
\newblock \showarticletitle{EasiEdge: A Novel Global Deep Neural Networks Pruning Method for Efficient Edge Computing}.
\newblock \bibinfo{journal}{\emph{IEEE Internet of Things Journal}} (\bibinfo{year}{2021}).
\newblock


\bibitem[Zhao and Luk(2018)]%
        {EfficientSP}
\bibfield{author}{\bibinfo{person}{Ruizhe Zhao} {and} \bibinfo{person}{Wayne W.~C. Luk}.} \bibinfo{year}{2018}\natexlab{}.
\newblock \showarticletitle{Efficient Structured Pruning and Architecture Searching for Group Convolution}.
\newblock \bibinfo{journal}{\emph{2019 IEEE/CVF International Conference on Computer Vision Workshop}}.
\newblock


\bibitem[Zoph and Le(2017)]%
        {NAS1}
\bibfield{author}{\bibinfo{person}{Barret Zoph} {and} \bibinfo{person}{Quoc Le}.} \bibinfo{year}{2017}\natexlab{}.
\newblock \showarticletitle{Neural Architecture Search with Reinforcement Learning}. In \bibinfo{booktitle}{\emph{International Conference on Learning Representations}}.
\newblock


\end{thebibliography}

\vspace{10pt}
\appendix
\noindent {\Large \textbf{Appendix}}

\section{Backpropagation is Memory Intensive}
Consider an \(N\)-layer CNN, denoted by \(F(\theta)\), where \(\theta\) belongs to the set \(\{\theta_{1},\theta_{2},....,\theta_{N}\}\) and \(\theta_{n}\) specifies the parameters in the \(n^{th}\) layer. Backpropagation-based training operates in two phases: the forward pass and the backward pass.

During the forward pass, the training input \(x\) traverses through the CNN \(F(\theta)\) with all intermediate activations \(a \in \{a_{1},a_{2},....,a_{N}\}\) being retained, thereby yielding the output \(\hat{y} = F(\theta ;x)\). This output \(\hat{y}\), in conjunction with the target output \(y\), is fed into a global loss function \(L(y,\hat{y})\) that represents the quality of the predicted output. 

Computation of the gradients \(\nabla L(\theta_{n})\) proceeds via the chain rule, as illustrated below:
\begin{equation}
\label{eq:global_gradient}
\frac{\partial L}{\partial \theta_{n}} = \frac{\partial L}{\partial a_n} \times \frac{\partial a_n}{\partial z_n} \times \frac{\partial z_n}{\partial \theta_{n}}
\end{equation}

The output of a neuron before activation is computed as:
\begin{equation}
\label{eq:pre_activation}
z_{n} = a_{n-1} \cdot \theta_{n}
\end{equation}

Differentiating this w.r.t. \(\theta_{n}\) gives:
\begin{equation}
\label{eq:pre_activation_diff}
\frac{\partial z_{n}}{\partial \theta_{n}} = a_{n-1}
\end{equation}

The output after the activation function, \( \sigma \), is given by:
\begin{equation}
\label{eq:activation}
a_n = \sigma(z_n)
\end{equation}

Considering the dependence on the activation function, the gradient of the loss with respect to the pre-activation output is:
\begin{equation}
\label{eq:activation_diff}
\frac{\partial a_n}{\partial z_n} = \sigma'(z_n)
\end{equation}

Substituting into Equation~\ref{eq:global_gradient} gives:
\begin{equation}
\label{eq:global_gradient_explicit}
\frac{\partial L}{\partial \theta_{n}} = \frac{\partial L}{\partial a_n} \times \sigma'(z_n) \times a_{n-1}
\end{equation}

After computing the gradients for the final layer, the backward pass begins to systematically propagate gradients from the last to the initial layer. The backward pass necessitates the preservation of all activations during the forward pass. This is because the gradient computation for layer \(n\) in the backward pass depends on the activations from the prior layer \((n-1)\) as shown in Equation~\ref{eq:global_gradient_explicit}. These gradients guide the parameter updates aimed at reducing the training loss.

\label{sec:appendix-BP}

\section{Theoretical Analysis}
The theoretical foundations of end-to-end CNN training using classic local learning have been presented in the literature~\cite{DGL}. The \accordion{} approach advances local learning by offering a strategy to segment the CNN into $M$ distinct blocks and train each block sequentially with unique batch sizes. Our analysis builds on existing work~\cite{DGL}, but our objective is to explore adaptive local learning underpinning \accordion{} that has not been considered elsewhere. Convergence is guaranteed under reasonable assumptions~\cite{Optimization_Methods}. Our starting point is the convergence dynamics of the first block. These are akin to patterns observed in smaller CNNs when trained using local learning~\cite{DGL}. 

\subsection{Convergence of the First Block}

Consider a block at iteration \( t \) with \( N \) layers. The parameters of the \( n^{th} \) layer (including the auxiliary parameters) are represented by \( \Psi_{n}^{t} \). While conventional end-to-end BP-based training maintains a fixed input distribution to the model, this is not the case with local learning. The input distribution to subsequent layers in local learning is time-varying, except for the first layer, which maintains a fixed input distribution. This time-varying nature of the input distribution is attributed to the fact that each layer, (except the first), is trained based on outputs generated from its preceding layer. As data is propagated and the layers are updated, the distribution of the outputs change over iterations.

The training batches can be represented as:
\begin{equation}
D_{j_{n}}^{t} \triangleq \{x^{t}_{j_{n}}, y_{j}\}_{j\leq J} 
\end{equation}
where \( x^{t}_{j_{n}} \) denotes the $j^{th}$ output batch of the \( n^{th} \) layer (input batch to the $(n+1)^{th}$ layer) at iteration \( t \), and its corresponding target batch \( y_{j}\) remains invariant over all iteration \( t \) and layers \( n \). Meanwhile, $J$ represents the cumulative number of training batches. Each training batch follows the density \( \rho_{n}^{t}(k) \). For layers with \( n > 1 \), the convergence density of its previous layer is \( \rho_{n-1}^{*}(k) \). The \textit{drift} of the preceding layer is described as:
\begin{equation}
s^{t}_{n-1} = \int | \rho_{n-1}^{t}(k) - \rho_{n-1}^{*}(k) | dk
\end{equation}

The \textit{drift} for the initial layer (\(n = 1\)) is zero, since its input distribution remains constant. The expected risk is:
\begin{equation}
\mathcal{L} = \mathop{\mathbb{E}_{\rho_{n-1}^{*}}[\ell(D_{j_{n-1}};\Psi_{n})]} 
\end{equation}
where \( \ell \) represents an invariably non-negative loss function. 

Consider the following assumptions. Assumption~1 to Assumption~3 are standard non-convex setting~\cite{Optimization_Methods} and Assumption~4 considers the time-varying input distribution~\cite{DGL}

\begin{itemize}[leftmargin=*]
    \item \textbf{Assumption 1} (L-smoothness and Differentiability): The gradient of \( \mathcal{L} \) is L-Lipschitz continuous. The learning rate of SGD is represented as \( \eta_{t} \) and for \( D_{j_{n-1}}^{t} \sim \rho_{n-1}^{t}(k) \). Our parameter sequence is:
    \begin{equation}
    \Psi_{n}^{t+1} = \Psi_{n}^{t} - \eta_{t}\nabla_{\Psi_{n}}\ell(D_{j_{n-1}}^{t};\Psi_{n}^{t})
    \end{equation}

    \item \textbf{Assumption 2} (Robbins-Monro Criterion): 
    \begin{equation}
    \sum_{t} \eta_{t} = \infty \quad \text{and} \quad \sum_{t} \eta_{t}^{2} < \infty
    \end{equation}

    \item \textbf{Assumption 3} (Bounded Variance): A boundary \( G > 0 \) exists such that for every \( t \) and \( \Psi_{n} \):
    \begin{equation}
    \mathbb{E}_{\rho_{n-1}^{t}}[||\nabla_{\Psi_{n}}\ell(D_{j_{n-1}};\Psi_{n})||^{2}] \leq G 
    \end{equation}

    \item \textbf{Assumption 4} (Stabilization of Prior Layer):
    \begin{equation}
    \sum_{t}s^{t}_{n-1} < \infty 
    \end{equation}
\end{itemize}

Given Assumption~3 and Assumption~4 and as already demonstrated in Lemma 4.1 presented in the literature~\cite{DGL}:
\begin{equation}
\mathbb{E}_{\rho_{n-1}^{*}}[||\nabla_{\Psi_{n}}\ell(D_{j_{n-1}};\Psi_{n})||^{2}] \leq G 
\end{equation}

Additionally, from Lemma 4.2 presented in the literature~\cite{DGL} and given Assumption~1, Assumption~3 and Assumption~4:
\begin{multline}
\mathbb{E}[\mathcal{L}(\Psi_{n}^{t+1})] \leq \mathbb{E}[\mathcal{L}(\Psi_{n}^{t})] \\ + \frac{LG}{2}\eta^{2}_{t} - \eta_{t}(\mathbb{E}[||\nabla\mathcal{L}(\Psi_{n}^{t})||^{2}] - \sqrt{2}Gs_{n-1}^{t})
\end{multline}

Furthermore, based on the above four assumptions and using Proposition 4.2 presented in the literature~\cite{DGL}, each term in the following equation converges.
\begin{multline}
\sum_{t = 0}^{T}\eta_{t}\mathbb{E}[||\nabla\mathcal{L}(\Psi_{n}^{t})||^{2}] \\ \leq \mathbb{E}[\mathcal{L}(\Psi_{n}^{0})] + G  \sum_{t = 0}^{T}\eta_{t}(\sqrt{2s_{n-1}^{t}} + \frac{L\eta_{t}}{2})
\end{multline}

Hence, the first block exhibits convergence. This is in alignment with the convergence criteria detailed in the literature~\cite{Optimization_Methods} and resembles end-to-end local learning of a small CNNs~\cite{DGL}. Upon completing the training of the first block, which comprises $N$ layers, the trained output batches (activations) from the final layer, denoted as \(D_{j_{N}}^{*} \triangleq \{x^{*}_{j_{N}}, y_{j}\}_{j<J}\), are saved to storage. 

\subsection{Convergence analysis of the Subsequent Blocks}
\subsubsection{Setting for the Second Block:}

Consider the second block at iteration \( t \), comprised of \( M \) layers. While the first layer of the second block can be perceived as the \( N+1 \) layer from the perspective of the previous block, we will use the index \( m \) to clarify layers within the second block, where \( m = 1, \dots, M \). The outputs (activations) from the final layer of the initial block become the inputs for this block.

Notably, this block is trained using a distinct batch size, leading to a different number of training batches, \( Q \). These training batches are represented as:
\begin{equation}
Z_{q_{m}}^{t} \triangleq \{x^{t}_{q_{m}}, y_{q}\}_{q<Q} 
\end{equation}

The input distribution for the layer \( m = 1 \) is static, reflecting the training dynamics of the first layer in the primary block. This is because the inputs for this layer are obtained from the outputs of the last layer of the preceding block, fetched directly from storage.

Each training batch conforms to the density \( \gamma_{m}^{t}(p) \). For layers with \( m > 1 \), we define the convergence density of its preceding layer as \( \gamma_{m-1}^{*}(p) \). This gives rise to the drift of the previous layer, defined as:
\begin{equation}
s^{t}_{m-1} = \int | \gamma_{m-1}^{t}(p) - \gamma_{m-1}^{*}(p) | dp
\end{equation}

The resultant expected risk is:
\begin{equation}
\mathcal{L} = \mathop{\mathbb{E}_{\gamma_{m-1}^{*}}[\ell(Z_{q_{m-1}};\Psi_{m})]} 
\end{equation}

\subsubsection{Analytical Similarity with the First Block:}

The analytical approach to understand the convergence of the second block shares resemblance with that of the first block. The distinction arises from the unique training dataset and batch size of the current block. The mathematical representations and assumptions applied to the first block can similarly be applied to the second block, establishing convergence on the second block. The analysis reinforces that the second block can be conceptualized as an independent CNN with its own dataset and batch size.

\subsubsection{Extension to Subsequent Blocks:}

Drawing upon the understanding of the first two blocks, it becomes clear that the analytical framework can be applied to all subsequent blocks within the CNN when utilizing adaptive local learning. The convergence of the entire CNN using \accordion{} can be understood as training several smaller, discrete, and independent CNNs via local learning~\cite{DGL}, with each having its own dataset and batch size.
\label{sec:appendix-convergence}

\section{Inference Throughput}
Table~\ref{tab:throughput_comparison} shows the inference throughput measured as images per second (images/s) of BP, classic LL and \accordion{} on different hardware platforms for VGG-16, VGG-19 and ResNet-18 on three different datasets. The speed-up \accordion{} achieves is shown alongside. BP and classic LL have the same throughput since the trained DNNs produced by these methods have the same number of parameters. 
\begin{table}[t]
\centering
\caption{Inference throughput (images per second; images/s) of the trained output DNN on different platforms, including the Tiny ImageNet dataset.}
\tiny
\begin{tabular}{|c|c|c|c|c|c|}
\hline
\multirow{2}{*}{\textbf{Platform}} & \multirow{2}{*}{\textbf{Dataset}} & \multirow{2}{*}{\textbf{Model}} & \multicolumn{2}{c|}{\textbf{images/s}} & \multirow{2}{*}{\textbf{Speedup}} \\
\cline{4-5}
 &  &  & \textbf{BP/classic LL} & \textbf{\accordion{}} & \\
\hline
\multirow{9}{*}{Pi4B} & \multirow{3}{*}{CIFAR-10} & VGG-16 & 6 & \textbf{19} & 3.17x \\
 &  & VGG-19 & 5 & \textbf{12} & 2.40x \\
 &  & ResNet-18 & 3 & \textbf{6} & 2.00x \\
\cline{2-6}
 & \multirow{3}{*}{CIFAR-100} & VGG-16 & 6 & \textbf{15} & 2.50x \\
 &  & VGG-19 & 5 & \textbf{12} & 2.40x \\
 &  & ResNet-18 & 3 & \textbf{5} & 1.67x \\
\cline{2-6}
 & \multirow{3}{*}{Tiny ImageNet} & VGG-16 & 6 & \textbf{15} & 2.50x \\
 &  & VGG-19 & 5 & \textbf{17} & 3.40x \\
 &  & ResNet-18 & 3 & \textbf{5} & 1.66x \\
\hline
\multirow{9}{*}{Jetson Nano} & \multirow{3}{*}{CIFAR-10} & VGG-16 & 213 & \textbf{706} & 3.31x \\
 &  & VGG-19 & 164 & \textbf{551} & 3.36x \\
 &  & ResNet-18 & 180 & \textbf{424} & 2.36x \\
\cline{2-6}
 & \multirow{3}{*}{CIFAR-100} & VGG-16 & 212 & \textbf{638} & 3.01x \\
 &  & VGG-19 & 160 & \textbf{554} & 3.46x \\
 &  & ResNet-18 & 181 & \textbf{364} & 2.01x \\
\cline{2-6}
 & \multirow{3}{*}{Tiny ImageNet} & VGG-16 & 213 & \textbf{632} & 2.96x \\
 &  & VGG-19 & 160 & \textbf{632} & 3.95x \\
 &  & ResNet-18 & 180 & \textbf{330} & 1.83x \\
\hline
\multirow{9}{*}{Xavier NX} & \multirow{3}{*}{CIFAR-10} & VGG-16 & 1278 & \textbf{3811} & 2.98x \\
 &  & VGG-19 & 1053 & \textbf{2590} & 2.46x \\
 &  & ResNet-18 & 870 & \textbf{1600} & 1.84x \\
\cline{2-6}
  & \multirow{3}{*}{CIFAR-100} & VGG-16 & 1272 & \textbf{3015} & 2.37x \\
 &  & VGG-19 & 1034 & \textbf{2642} & 2.56x \\
 &  & ResNet-18 & 899 & \textbf{1481} & 1.65x \\
\cline{2-6}
 & \multirow{3}{*}{Tiny ImageNet} & VGG-16 & 1270 & \textbf{2997} & 2.35x \\
 &  & VGG-19 & 842 & \textbf{2960} & 3.51x \\
 &  & ResNet-18 & 885 & \textbf{1430} & 1.61x \\
\hline
\multirow{9}{*}{AGX Orin} & \multirow{3}{*}{CIFAR-10} & VGG-16 & 3706 & \textbf{10995} & 2.97x \\
 &  & VGG-19 & 3512 & \textbf{8271} & 2.36x \\
 &  & ResNet-18 & 2761 & \textbf{5779} & 2.09x \\
\cline{2-6}
 & \multirow{3}{*}{CIFAR-100} & VGG-16 & 3600 & \textbf{9746} & 2.71x \\
 &  & VGG-19 & 3442 & \textbf{8126} & 2.36x \\
 &  & ResNet-18 & 2578 & \textbf{4766} & 1.85x \\
\cline{2-6}
 & \multirow{3}{*}{Tiny ImageNet} & VGG-16 & 3450 & \textbf{9688} & 2.8x \\
 &  & VGG-19 & 3320 & \textbf{9647} & 2.9x \\
 &  & ResNet-18 & 2392 & \textbf{4391} & 1.83x \\
\hline
\end{tabular}
\label{tab:throughput_comparison}
\end{table}

\label{sec:appendix-inferencethroughput}

\end{document}